\documentclass[letterpaper]{article} % DO NOT CHANGE THIS
\usepackage{aaai24}  % DO NOT CHANGE THIS
\usepackage{times}  % DO NOT CHANGE THIS
\usepackage{helvet}  % DO NOT CHANGE THIS
\usepackage{courier}  % DO NOT CHANGE THIS
\usepackage[hyphens]{url}  % DO NOT CHANGE THIS
\usepackage{graphicx} % DO NOT CHANGE THIS
\usepackage{natbib}  % DO NOT CHANGE THIS AND DO NOT ADD ANY OPTIONS TO IT
\usepackage{caption} % DO NOT CHANGE THIS AND DO NOT ADD ANY OPTIONS TO IT
\usepackage{algorithm}

\usepackage{amsfonts, amsmath, amssymb, amsthm}
\usepackage{xspace}

\newcommand{\app}[0]{Appendix}

\newif\ifappendix
 \global\appendixtrue
\newcommand{\joint}[1]{\vec{#1}}
\newcommand{\cg}{\mathcal{CG}}
\newcommand{\ucb}{\mathcal{UCB}}

\newcommand{\given}{\,|\,}

\newcommand{\fgstep}{s', \; \vec{o}, \; r \sim \mathcal{G}(s, \vec{a})}

\newcommand{\tuple}[1]{\langle #1 \rangle}

\usepackage{silence}
\usepackage{algorithm}
\usepackage[noend]{algpseudocode}
\usepackage{xcolor}

\definecolor{gray}{cmyk}{0,0,0,0.8}

\RequirePackage{xcolor}
\definecolor{blue_l}{RGB}{166,206,227}
\definecolor{blue_d}{RGB}{ 31,120,180}
\definecolor{green_l}{RGB}{178,223,138}
\definecolor{green_d}{RGB}{ 51,160, 44}
\definecolor{red_l}{RGB}{251,154,153}
\definecolor{red_d}{RGB}{227, 26, 28}
\definecolor{orange_l}{RGB}{253,191,111}
\definecolor{orange_d}{RGB}{255,127,  0}
\definecolor{purple_l}{RGB}{202,178,214}
\definecolor{purple_d}{RGB}{106, 61,154}
\definecolor{brown_l}{RGB}{255,255,153}
\definecolor{brown_d}{RGB}{177, 89, 40}
\definecolor{yellow_l}{RGB}{255,255,153}
\definecolor{yellow_d}{RGB}{177, 89, 40}

\RequirePackage{xargs}       % Use more than one optional parameter in a new commands

\RequirePackage[%
  backgroundcolor=white,
  textsize=tiny,
]{todonotes}
  \newcommandx{\mg}[2][1=]{\todo[color=blue_l,   linecolor=blue_l,   bordercolor=blue_d,    #1]{\textbf{MG:} #2}}
  \newcommandx{\ts}[2][1=]{\todo[color=purple_l, linecolor=purple_l, bordercolor=purple_d,  #1]{\textbf{TS:} #2}}
  \newcommandx{\sj}[2][1=]{\todo[color=green_l,  linecolor=green_l,  bordercolor=green_d,   #1]{\textbf{SJ:} #2}}
  \newcommandx{\nj}[2][1=]{\todo[color=orange_l, linecolor=orange_l, bordercolor=orange_d,  #1]{\textbf{NJ:} #2}}

\usepackage{silence}
\newtheorem{theorem}{Theorem}%[section]
\newtheorem{definition}{Definition}
\newtheorem{assumption}{Assumption}

\usepackage{cleveref}

\crefname{equation}{Eq.}{Eq.}
\crefname{pluralequation}{Eqs.}{Eqs.}

\crefname{algorithm}{Algorithm}{Algorithm}

\crefname{figure}{Fig.}{Fig.}
\crefname{pluralfigure}{Figs.}{Figs.}

\crefname{section}{Sect.}{Sect.}
\crefname{pluralsection}{Sects.}{Sects.}

\crefname{table}{Table}{Table}
\crefname{pluraltable}{Tables}{Tables}

\crefname{theorem}{Theorem}{Theorems}
\crefname{pluraltheorem}{Theorems}{Theorems}

\crefname{definition}{Def.}{Def.}
\crefname{pluraldefinition}{Defs.}{Defs.}

\crefname{lemma}{Lemma}{Lemma}
\crefname{plurallemma}{Lemmas}{Lemmas}

\crefname{example}{Example}{Example}
\crefname{pluralexample}{Examples}{Examples}

\crefname{assumption}{Assumption}{Assumption}
\crefname{pluralassumption}{Assumptions}{Assumptions}

\crefname{remark}{Remark}{Remark}
\crefname{pluralremark}{Remarks}{Remarks}

\crefname{appendix}{\app}{\app}
\crefname{pluralappendix}{Apps.}{Apps.}

\usepackage{caption}
\usepackage{subcaption}

\DeclareMathOperator*{\argmax}{arg\,max}

\newcommand{\ie}{i.\,e.\@\xspace}

\usepackage{booktabs}
\usepackage{multirow}
\usepackage{comment}
\usepackage{mdframed}
\usepackage{fontawesome5}
\usepackage{makecell}
\usepackage{colortbl}
\usepackage{nicefrac}
\usepackage{mathtools}

\newcommand{\introterm}[1]{\textbf{#1}}
\newcommand{\introalgo}[1]{#1}

\title{Factored Online Planning in Many-Agent POMDPs}
\author{
    Maris F.L. Galesloot\textsuperscript{\rm 1},
    Thiago D. Sim\~ao\textsuperscript{\rm 2},
    Sebastian Junges\textsuperscript{\rm 1},
    Nils Jansen\textsuperscript{\rm 1,\rm 3}
}
\affiliations{
    \textsuperscript{\rm 1}Radboud University Nijmegen, The Netherlands \\
    \textsuperscript{\rm 2}Eindhoven University of Technology, The Netherlands \\
    \textsuperscript{\rm 3}Ruhr-University Bochum, Germany \\
    maris.galesloot@ru.nl, t.simao@tue.nl, sebastian.junges@ru.nl, n.jansen@rub.de
    
}

\begin{document}

\maketitle

\begin{abstract}
    In centralized multi-agent systems, often modeled as multi-agent partially observable Markov decision processes~(MPOMDPs), the action and observation spaces grow exponentially with the number of agents, making the value and belief estimation of single-agent online planning ineffective.
    Prior work partially tackles \textit{value estimation} by exploiting the inherent structure of multi-agent settings via so-called coordination graphs.
    Additionally, \textit{belief estimation} methods have been improved by incorporating the likelihood of observations
    into the approximation.
    However, the challenges of value estimation and belief estimation have 
    only been tackled individually,
    which prevents existing methods from scaling to settings with many agents.
    Therefore, we address these challenges simultaneously.
    First, we introduce \textit{weighted particle filtering} 
    to a sample-based online planner for 
    MPOMDPs. Second, we present a scalable approximation of the belief.
    Third, we bring an approach that exploits the typical locality of agent interactions  
    to novel online planning algorithms for MPOMDPs operating on a so-called sparse particle filter tree.
    Our experimental evaluation against several state-of-the-art baselines shows that our methods (1) are competitive in settings with only a few agents and (2) improve over the baselines in the presence of many agents.
\end{abstract}

\begin{table*}[tb]
\centering
\begin{tabular}{@{}lllll@{}}
\toprule
                       % &            &  \multicolumn{2}{c}{\textbf{Value Function}} \\  \cmidrule(l){3-4} 
\multirow{2}{*}{\textbf{Belief Estimation}}                       & \multirow{2}{*}{\textbf{Simulation}}           &  \multicolumn{1}{l}{\textbf{POMDP Value Estimation}} & \multicolumn{2}{l}{\textbf{MPOMDP Value Estimation}}     \\ 
\cmidrule(l){3-5} & & \makecell[l]{\textrm{Single-Agent}} & \makecell[l]{\textrm{Factored Statistics}} & \makecell[l]{\textrm{Factored Trees}} \\
\cmidrule[.7pt](r){1-2}
\cmidrule[.7pt](lr){3-3}\cmidrule[.7pt](lr){4-4}\cmidrule[.7pt](l){5-5}
\multirow{2}{*}{Unweighted} & \multirow{2}{*}{Single States}  & \makecell[l]{\introalgo{POMCP}}  & \makecell[l]{\introalgo{FS-POMCP}} & \makecell[l]{\introalgo{FT-POMCP}} \\
& & \cite{DBLP:conf/nips/SilverV10}& \multicolumn{2}{c}{\cite{DBLP:conf/aaai/AmatoO15}} \\
\cmidrule{1-5} 
                   % &  Belief estimates    &  -           &  - \\
%\cmidrule(lr){1-1} \cmidrule(l){2-2}
\multirow{4}{*}{Weighted} & \multirow{2}{*}{Single States} &  \makecell[l]{\introalgo{W-POMCP} / POMCPOW} &\makecell[l]{\introalgo{FS-W-POMCP}} & \makecell[l]{\introalgo{FT-W-POMCP}}\\
& & \citep{DBLP:conf/aips/SunbergK18} & \multicolumn{2}{c}{\cref{sec:scalable_pf} (\textit{new})}
\\ \cmidrule{2-5} 

&  \multirow{2}{*}{Belief Estimates} & \makecell[l]{\introalgo{Sparse-PFT}} & \makecell[l]{\introalgo{FS-PFT}} & \makecell[l]{ \introalgo{FT-PFT}} \\
& & \makecell[l]{\cite{lim2023optimality}} & \multicolumn{2}{c}{\cref{sec:factored_pft} (\textit{new})} \\
\bottomrule
\end{tabular}
\caption{Our algorithms and state-of-the-art MCTS methods in the (continuous) POMDP and MPOMDP literature.}
\label{tab:intro}
\end{table*}

\section{Introduction}
Planning problems with multiple agents, such as teams of mobile robots \citep{DBLP:conf/cdc/AhmadiSBA19} or autonomous surveillance systems \citep{DBLP:journals/ram/WitwickiCMCMLV17}, can be modeled by multi-agent partially observable Markov decision processes~\citep[MPOMDPs,][]{DBLP:conf/nips/MessiasSL11}.
% Why is this hard
These formal models exhibit sets of (local) observations and actions for each agent that can be shared with a central controller.
This controller then makes decisions among the joint actions of all agents.
Computationally, the main challenge is that the spaces of joint action and observations grow exponentially with the number of agents~\citep{DBLP:journals/jair/PynadathT02}.
Moreover, as the controller only partially observes the system state,
it must base its decisions on the history of previous joint actions and observations.

% \noindent
Online algorithms, such as those based on \textit{Monte Carlo tree search} \citep[MCTS,][]{DBLP:journals/tciaig/BrownePWLCRTPSC12}, are a common way to tackle large planning problems. 
% A common way to tackle large planning problems is to employ online algorithms, such as \textit{Monte Carlo tree search}~\cite[MCTS,][]{DBLP:journals/tciaig/BrownePWLCRTPSC12}. 
These algorithms search for local solutions in the most promising regions of the search space~\cite{DBLP:conf/ecml/KocsisS06}.
In particular, \textit{partially observable Monte Carlo planning} \citep[POMCP,][]{DBLP:conf/nips/SilverV10} derives Monte Carlo estimates in the form of (1)~an approximation of the value function and (2)~distributions~(beliefs) over states by generating sample trajectories from the simulation of single state particles.
However, a naive application of (single-agent) online planning is ill-equipped to handle the high-dimensional MPOMDP setting.
First, due to the many actions that must be explored during simulations, the \introterm{value estimation} may suffer from high variance. 
Second, the chance of mismatch between simulated and actual observations is high for large observation spaces, lowering the quality of the approximation in the \introterm{belief estimates}~\citep{DBLP:conf/aips/SunbergK18}.

% \noindent
To address the challenge of {value estimation}, one can exploit the typical locality of interactions between the agents, captured by so-called \introterm{coordination graphs}~\citep{DBLP:conf/icml/GuestrinLP02}.
In particular, \citet{DBLP:conf/aaai/AmatoO15} estimate the action value for subsets of agents instead of all agents based on such graphs.
The main concepts are to (1) factorize the value estimates over the action space of subsets of agents in the \introterm{factored statistics} variant and to (2) factorize both the action and the observation space in the \introterm{factored trees} variant.
These factorizations are key to achieving good performance in settings with many agents.
The challenge of belief estimation is also a prevalent issue in single-agent continuous settings.
From the likelihood of sampled observations, importance sampling weights are added to the Monte Carlo estimates of the beliefs~\citep{DBLP:conf/nips/Thrun99}.
Such {weighted beliefs} are also used in single-agent online planners that simulate \introterm{weighted belief estimates} instead of {single states}~\citep{DBLP:conf/icml/0007T20,DBLP:conf/ijcai/LimTS20}.
By simulating belief estimates, these algorithms operate on the set of possible beliefs of the agents, which makes the branching factor insensitive to the number of observations.
A particularly effective algorithm is the so-called \textit{sparse particle filter tree}~\citep[Sparse-PFT,][]{lim2023optimality}, which only searches for local solutions in the set of reachable belief estimates.

% \noindent
To the best of our knowledge, these solutions to value and belief estimation have only been studied independently, and weighted belief estimates have not yet been explored in the online MPOMDP planning setting.
Therefore, no method can scale to problems with many agents, and we must tackle both challenges simultaneously.
We present multi-agent online planning algorithms that exploit adequate approximations of both the {value} and the {belief} and thereby can scale to many agents.
In particular, we integrate factored value estimation and weighted belief estimation.
\cref{tab:intro} positions the new algorithms with respect to the existing solutions in the MCTS literature.
First, \textit{we add weighted belief estimation to POMCP variants}, namely \introalgo{W-POMCP}, and combine this algorithm with \textit{factored statistics} in \introalgo{FS-W-POMCP}~(\cref{sec:wpf}). Furthermore, we design a weighted belief approximation that is compatible with \textit{factored trees} in \introalgo{FT-W-POMCP}~(\cref{sec:scalable_pf_method}). 
Then, we \textit{introduce two novel variants} of the Sparse-PFT algorithm that exploit coordination graphs similarly to \citet{DBLP:conf/aaai/AmatoO15}, namely \introalgo{FS-PFT} and \introalgo{FT-PFT}~(\cref{sec:fsandftpft}).
The empirical evaluation (1) shows the improvement of integrating weighted particle filtering in POMCP, and (2) demonstrates the effect of value factorization via coordination graphs in Sparse-PFT.
It also shows that exploiting local interactions between agents is beneficial in environments
where the way the agents interact may change over time, \ie, there is no naturally sparse graph.
\ifappendix
This manuscript is an extended version of the article containing an Appendix with additional details.
\else
The extended version of this article~\citep{galesloot2023factored} contains an Appendix with additional details.
\fi

\paragraph{Contributions.}
We present four novel algorithms for many-agent online planning that can scale both value and belief estimations to problems with many agents.
Our empirical evaluation, using the nine algorithms in \cref{tab:intro}, shows that our variants improve over the state-of-the-art.
For example, in the environments \textsc{FireFightingGraph} and Multi-Agent \textsc{RockSample}, we scale up to instances with 64 and 6 agents instead of 10 and 2 agents, respectively.

%%%%%%%%%%%%%%%%%%%%
% BACKGROUND START %
%%%%%%%%%%%%%%%%%%%%

\section{Online Planning in MPOMDPs}
\label{sec:problem}
The set of all distributions over the finite set $X$ is $\Delta(X)$.
\paragraph{MPOMDPs.}
We study online planning in centralized multi-agent systems that are modeled as MPOMDPs.
Intuitively, agents encounter individual observations but can share those via immediate and noiseless broadcast communication, which allows a centralized control paradigm. 
\begin{definition}[\textbf{MPOMDP}]
\label{def:mpomdp}
An \emph{MPOMDP} is a tuple $\mathcal{M} = \tuple{\mathcal{I}, \mathcal{S}, b_0, \mathcal{A}, \mathcal{T}, \mathcal{R}, \Omega, \mathcal{O}, \gamma}$, with
    the finite set $\mathcal{I}$ of $n$ \emph{agents},
    the finite set $\mathcal{S}$ of \emph{states}, an \emph{initial state distribution} $b_0 \in \Delta(\mathcal{S})$, the set 
    $\mathcal{A} = \bigtimes_{i\in\mathcal{I}}\mathcal{A}_i$ of \emph{joint actions}, composed of the finite sets $\mathcal{A}_i$ of \emph{actions} for each agent $i \in \mathcal{I}$,
   the \emph{transition function} $\mathcal{T} \colon \mathcal{S} \times \mathcal{A} \rightarrow \Delta(\mathcal{S})$  such that $\mathcal{T}(s'\given s, \joint{a}) = \Pr(s' \given s, \joint{a})$ is the probability of a new state $s'$ given the previous state $s$ and joint action $\joint{a}$,
   the \emph{reward function} $\mathcal{R}\colon \mathcal{S} \times \mathcal{A} \rightarrow \mathbb{R}$  such that $\mathcal{R}(s, \joint{a})$ is the \emph{reward} given state $s$ and joint action $\joint{a}$, 
    the set $\Omega = \bigtimes_{i\in \mathcal{I}}\Omega_i$ of \emph{joint observations} composed by the finite sets $\Omega_i$ of \emph{observations}  for each agent $i \in \mathcal{I}$,
   the \emph{observation function} $\mathcal{O} \colon \mathcal{S} \times \mathcal{A} \rightarrow \Delta(\Omega)$ such that $\mathcal{O}(\joint{o} \given s',\joint{a}) = \Pr(\joint{o} \given s', \joint{a})$ specifies the probability of observing joint observation $\joint{o}$ in the state $s'$ given joint action~$\joint{a}$, and the \emph{discount factor}
    $\gamma \in [0,1)$.
\end{definition}
\noindent
MPOMDPs generalize POMDPs~\citep{DBLP:journals/ai/KaelblingLC98}, which are MPOMDPs with a single agent. An MPOMDP can be treated as a POMDP by ignoring the agent-wise factorization in the action and observation space.
\paragraph{Objective.}
The \textit{return} $R_t = \sum_{t'=t}^{\infty} \gamma^{t'-t} \mathcal{R}(s_{t'}, a_{t'})$ is the infinite-horizon discounted sum of reward from time $t\in\mathbb{N}$.
 % the expected return from $t{=}0$.
An observable \textit{history} $\joint{h}_t = (\joint{o}_1, \joint{a}_1, \joint{o}_2, \ldots, \joint{a}_{t-1}, \joint{o}_t)$ is a sequence of joint observations and joint actions. 
Policies determine the action choices. 
Optimal policies $\pi \colon \Delta(S) \rightarrow \mathcal{A}$ for MPOMDPs map the belief $b_t\in \Delta(\mathcal{S})$ to joint actions.
The belief $b_t$ is a \textit{sufficient statistic} \citep{DBLP:journals/ai/KaelblingLC98} for the history $h_t$, and resembles the state distribution $b_t(s) = \Pr(s_t \mid h_t, b_0)$ at time $t$, with $b_0$ from~$\mathcal{M}$.
Beliefs can be updated $b_{t+1} = \upsilon(b_t, \joint{o}_{t+1}, \joint{a}_{t})$ by $\upsilon$ from $\mathcal{T}$ and $\mathcal{O}$, using Bayes' theorem \citep{DBLP:books/sp/12/Spaan12}.
Our aim is to maximize the \textit{joint Q-value} of a belief $b$, which is the \textit{expected return} under a policy $\pi$ given action $\joint{a}$ and belief $b$ at time $t$, and $a_{t'} = \pi(b_t)$ for subsequent $t'>t$:
\begin{equation}
    Q^\pi(b, \joint{a}) 
    = \mathbb{E}_{\pi}
    \left[
    R_t
    \mid b_t{=}b, \joint{a}_t{=}\joint{a}
    , a_{t'>t} = \pi(b_{t'})
    \right].
\end{equation}
\paragraph{Online planning.}
Online search-based planners interleave planning and execution.
They perform a forward search in the set of beliefs reachable from the current belief, incrementally building a look-ahead tree known as a search tree. 
Monte Carlo planners typically do so with a generative interface $\mathcal{G} \colon \mathcal{S} \times \mathcal{A} \rightarrow \mathcal{S} \times \Omega \times \mathbb{R}$ of the model $\mathcal{M}$, \ie, a \textit{simulator}~\citep{DBLP:journals/ml/KearnsMN02}, with $s, \joint{a} \mapsto s', \joint{o}, r$.
After 
searching from $b$, the planner executes a selected action $\joint{a}$, receives an observation $\joint{o}$. Then, it updates its belief $b' = \upsilon(b, \joint{a}, \joint{o})$, before it starts searching from $b'$.

\paragraph{MCTS.}\label{sec:mcts-pomcp}
Contemporary MCTS methods are based on the \textit{upper confidence trees}~\citep[UCT,][]{DBLP:conf/ecml/KocsisS06} algorithm. In particular,
partially observable UCT (PO-UCT) is the search algorithm that underlies POMCP. It plans with a look-ahead search tree comprised of paths of action and observation nodes.
It samples states from the current belief $b$, and for each state, it expands a trajectory of actions and observations
using $\mathcal{G}$ 
until 
it reaches a new node in the tree. Then, a (random) \textit{rollout} estimates the value \citep{DBLP:conf/ecml/KocsisS06}.
The trajectory is used to update a set of statistics for history $\joint{h}$ that includes visit counts $N(\joint{h})$ and $n(\joint{h}, \joint{a})$ in the observation and action nodes, respectively.
Additionally, the trajectory updates estimates of the Q-values $Q(\joint{h}, \joint{a})$ of the action nodes by a running average of the return.
The upper confidence bound~\citep[UCB1,][]{DBLP:journals/ml/AuerCF02} algorithm decides the most promising actions during the search, balancing exploration and exploitation. It is computed from (an estimate of) the number $N^{*}$ of visits to the observation node, as well as the number of visits~$n$ and the value $Q^{*}$ of the action node by: 

\begin{equation}\label{eq:ucb}\textstyle
\ucb(Q^{*}, N^{*}, n^{*}) = Q^{*} + c\cdot \sqrt{\frac{\log(N^{*}+1)}{(n^{*}+1)}},
\end{equation}
where $c$ is an exploration constant. During search, in some history $\joint{h}$,
PO-UCT chooses actions via:
$
\argmax_{\joint{a}}  ~\ucb(Q(\joint{h}, \joint{a}), N(\joint{h}), n(\joint{h}, \joint{a})).
$

\paragraph{A separation of beliefs.} POMCP is the extension of PO-UCT that gradually builds up beliefs $B(\joint{h})$ consisting of simulated states $s\in\mathcal{S}$, \ie, \textit{particles}, in the observation nodes for history $\joint{h}$, representing a Monte Carlo estimate of the belief $b$. However, the number of particles in each observation node depends on how often a history has been recorded during forward simulation and might diminish over time due to a lack of diversity in the set of particles. 
POMCP requires domain-specific particle reinvigorating techniques to mitigate this.
Instead, we separate the concerns of an \textit{online} belief \textit{inside the search tree} that is used to estimate $Q$, 
and the \textit{offline} belief $\overline{b}$ that represents the current belief over states.
Following related works, we write $B(\joint{h})$ for POMCP's online belief and $\tilde{b}$ for Sparse-PFT's weighted online belief.
\begin{mdframed}
    \textbf{Problem statement:} Given a 
    an MPOMDP $\mathcal{M}$, how do we, at each time step $t \in \mathbb{N}$, both (1) efficiently search for a joint action $\joint{a}_t$ given the current belief estimate $\overline{b}_t$, and (2) effectively find a good belief estimate 
    $\overline{b}_{t+1}$ from $\overline{b}_{t}, \joint{a}_t$ and the received observation~$\joint{o}_{t+1}$.
\end{mdframed}

%%%%%%%%%%%%%%%%%%%
% BACKGROUND DONE %
%%%%%%%%%%%%%%%%%%%

\section{Using Structure in Multi-Agent POMDPs}
\label{sec:structurepomdps}
In this section, we introduce \textit{coordination graphs} to decompose the objective into local sub-problems. In particular, we recap prior work by \citet{DBLP:conf/aaai/AmatoO15}.

\subsection{Coordination Graphs}\label{sec:moe}
A \emph{coordination graph} \citep[CG,][]{DBLP:conf/aaai/GuestrinVK02,DBLP:series/sbis/OliehoekA16} is an undirected graph $(\mathcal{V}, \mathcal{E})$ that represents the local interactions between agents.
Each vertex $v \in \mathcal{V}$ corresponds to an agent ($\mathcal{V} \equiv \mathcal{I}$), and each edge $(i, j) \in \mathcal{E}$ indicates that agents $i \in \mathcal{V}$ and $j \in \mathcal{V}$ interact locally.
For an edge $e \in \mathcal{E}$, we define the \emph{local action} $\joint{a}_e$ and \emph{local observations} $\joint{o}_e$, which  range over the product of the individual agent action $\mathcal{A}_{e}=\mathcal{A}_i \times \mathcal{A}_j$ and observation spaces $\Omega_{e}= \Omega_i \times \Omega_j$, with $e= (i,j)$. 
For three agents $\mathcal{V} = \{1,2,3\}$ connected by a line $\mathcal{E} = \{e_1, e_2\}$, with $e_1 = (1, 2)$ and $e_2 = (2, 3)$, we have $\joint{a} = \{a_1, a_2, a_3\}$, thus $\joint{a}_{e_1} = \{a_1, a_2\}$ and $\joint{a}_{e_2} = \{a_2, a_3\}$, respectively.
To find the Q-value for some history $\joint{h}$ (or equivalent belief $b$) based on the local actions, we define a local payoff function $Q_e(\joint{h}, \joint{a}_e)$ for each edge $e \in \mathcal{E}$, where $\joint{a}_e\in\mathcal{A}_e$ is the projection of $\joint{a}$ to the agents in the edge. Then, $Q(\joint{h}, \joint{a}) \approx \sum_e Q_e(\joint{h}, \joint{a}_e)$.
Instead of finding $Q_e(\joint{h}, \joint{a}_e)$, we maintain local predictions $\hat{Q}_e(\joint{h}, \joint{a}_e) = \mathbb{E} \left[ Q(\joint{h}, \joint{a}) \mid \joint{a}_e \right]$ of the joint Q-value. 
% To estimate the joint Q-value, 
A \emph{mixture of experts}~(MoE) combines these local estimates of $Q$:
\begin{equation}\label{eq:moe}\textstyle
    Q(\joint{h}, \joint{a}) \approx \hat{Q}(\joint{h}, \joint{a}) = \sum_e \omega_e \hat{Q}_e(\joint{h}, \joint{a}_e),
\end{equation}
where $\omega_e \geq 0$ is the weight for edge $e$, s.t. $\sum_{e \in \mathcal{E}}\omega_e = 1$.
We assume uniform mixture weights $\omega_e = \nicefrac{1}{|\mathcal{E}|}$ throughout the remainder of the paper.
We pick the estimated maximizing joint action $\joint{a}^{\#} \approx \joint{a}^{*}$ over the sum of local estimates of $Q$:
\begin{equation}\label{eq:max-over-moe}\textstyle
    \joint{a}^{\#} = \argmax_{\joint{a}} \sum_e \omega_e \hat{Q}_e(\joint{h}, \joint{a}_e).
\end{equation}
We thus aim to find local actions (for the edges of the graph) that maximize the estimated joint value function.  
Notice that when finding $\joint{a}^\#$, any agent $i$ might belong to multiple edges, and therefore agent $i$ must be assigned the same action $a^\#_i \in \mathcal{A}_i$ in all edges $e \in \mathcal{E}$ where $i \in e$.
We can compute the maximum with graphical inference algorithms, such as Variable Elimination~(VE) and Max-Plus~(MP) \citep{DBLP:conf/smc/VlassisEK04}.
\ifappendix
    \cref{sec:action_selection}
\else
    \app~D
\fi
provides an overview.
% of these algorithms. 

\subsection{Factored-Value POMCP}\label{sec:fv_pomcp}
\emph{Factored-value} POMCP \citep{DBLP:conf/aaai/AmatoO15} consists of two techniques that exploit the structure of a CG to scale POMCP to problems with large action and observation spaces.
Next, we outline how these techniques factor the action space to introduce statistics for each edge $e \in \mathcal{E}$ for computing the UCB1 value of the local joint action space~$\joint{a}_e$. Both these algorithms require an inference algorithm to compute \cref{eq:max-over-moe} during and after the simulations.

\paragraph{Factored statistics (FS-POMCP).}
FS-POMCP uses the structure of MPOMDPs and stores the statistics $Q, N, n$ in a factorized manner. This adaption is more space-efficient and also allows for improved action selection in large action spaces by maximizing over the factored Q-functions. 
More precisely, the tree structure in FS-POMCP remains the same as in POMCP, representing the history $\joint{h}$ with associated visit counts $N(\joint{h})$ and particles $B(\joint{h})$ in the observation nodes. The action nodes maintain a set of statistics $Q_e(\joint{h}, \joint{a}_e)$, $N(\joint{h}, \joint{a}_e)$  for each edge $e \in \mathcal{E}$, independently.
Thus, the MoE optimization from \cref{eq:moe} is applied directly in each action node of the search tree.
This improves over POMCP as the combination of local action spaces $\joint{a}_e$ of each edge $e \in \mathcal{E}$ is smaller than the joint action space $\joint{a}$. 
During search, the action $\joint{a}^\#$ is selected by maximizing over the UCB1 values (\cref{eq:ucb}) of the local Q-functions:
$
    \argmax_{\joint{a}^\#} \sum_{e\in\mathcal{E}} \ucb(Q_e(\joint{h}, \joint{a}_e), N(\joint{h}), N(\joint{h}, \joint{a}_e)).
$
\paragraph{Factored trees (FT-POMCP).}
FT-POMCP constructs a tree for every edge $e$. This tree represents the factored histories $\joint{h}_e$, which consists of a sequence of factored actions and observations $\joint{h}_{e,t} = (\joint{a}_{e,0}, \joint{o}_{e,1}, \ldots, \joint{a}_{e,t-1}, \joint{o}_{e,t})$. 
This further reduces the scope of $Q_e(\joint{h}, \joint{a}_e)$ to $Q_e(\joint{h}_e, \joint{a}_e)$ by introducing an expert for every $\joint{h}_e, \joint{a}_e$ pair.
In each tree, the action nodes maintain statistics $Q_e(\joint{h}_e, \joint{a}_e)$, and $N(\joint{h}_e, \joint{a}_e)$, and the observation nodes maintain $N(\joint{h}_e)$ and $B(\joint{h}_e)$ according to the factored history $\joint{h}_e$. 
During search, we again maximize with respect to the UCB1 value (using \cref{eq:ucb}) of local Q-functions:
$
\argmax_{\joint{a}^\#} \sum_{e\in \mathcal{E}} \ucb(Q_e(\joint{h}_e, \joint{a}_e), N(\joint{h}_e), N(\joint{h}_e, \joint{a}_e)).
$

%%%%%%%%%%%%%%%%%%%%%%%%%%%%
% Coordination graphs done %
%%%%%%%%%%%%%%%%%%%%%%%%%%%%

\section{Scalable Particle Filtering}\label{sec:scalable_pf}
In MPOMDPs with large state spaces, repeated execution of the Bayesian belief update is intractable as each update requires $\mathcal{O}(|\mathcal{S}|^2)$ computations. 
We represent this belief with \textit{weighted particle filters} to ensure scalability. The following subsection introduces how we incorporate these filters in W-POMCP and FS-W-POMCP. The second subsection introduces our method that uses the structure of a coordination graph to condition the belief on a local part of the observation space, which we apply to FT-W-POMCP and FT-PFT.

\paragraph{Particle filtering.}
Particle filtering~\citep{DBLP:books/daglib/0014221} represents the belief by sequential Monte Carlo approximations, alleviating the bottleneck of the belief update. 
In an \textit{unweighted} filter, the belief approximation is a set $\overline{b}=\{(s^{(k)})\}_{k=1}^K$, with $s^{(k)}\in\mathcal{S}$, and is updated using rejection sampling on the real observation $\joint{o}$; $\overline{b}' = \{{s'}^{(k)}\colon \joint{o} = \joint{o}^{\;(k)}\}$, where ${s'}^{(k)}$, $\joint{o}^{\;(k)}$ are generated from $s^{(k)},\joint{a}$ by $\mathcal{G}$~\citep{kochenderfer_decision_2015}. 
POMCP implicitly uses an unweighted particle filter by using the online particle belief $B(\joint{h})$ in the observation nodes to represent $\overline{b}$.
\subsection{Weighted Particle Filtering}\label{sec:wpf}
\textit{Weighted particle filters} approximate the belief by a weighted set of $K$ particles $\overline{b}=\{(s^{(k)}, w^{(k)})\}_{k=1}^K$, where $s^{(k)}\in\mathcal{S}$ is a state in the filter and $w^{(k)}\in\mathbb{R}^+$ the associated weight. 
We update beliefs in weighted filters with \textit{importance sampling} as in the \textit{bootstrapped particle filter} \citep{Gordon_Salmond_Smith_1993}.
In the bootstrapped particle filter, the proposal distribution is the transition function ${s'}^{(k)} \sim \mathcal{T}(\cdot \given s^{(k)}, \joint{a})$, and the importance weights are computed from the observation function ${w'}^{(k)} \propto w^{(k)} \mathcal{O}(\joint{o}\given {s'}^{(k)}, \joint{a})$.
Additionally, the posterior belief is re-sampled at every time step to alleviate sample degeneracy, after which the weights are set to $\nicefrac{1}{K}$, which is known as \emph{sequential importance re-sampling}~(SIR).
We decide whether to re-sample in our SIR filter by comparing the estimated \textit{effective sample size} (ESS) of the particle filters with respect to the number of particles~\citep{DBLP:journals/jstsp/SeptierP16}. The $\text{ESS}(\overline{b}) \approx (\sum_{k=1}^K (w^{(k)})^2)^{-1}$ quantifies weight disparity, which is an indicator for sample degeneracy.
The \textit{likelihood} $\mathcal{L}$ of a belief update in a SIR filter represents the probability of the new belief given the observation, action, and previous belief. It is a statistic on the quality of the approximate belief update~\citep{DBLP:conf/atal/KattOA19}. It is computed from the sum of all updated weights multiplied by the previous likelihood $\mathcal{L}(\overline{b}') = \sum_{k} {w'}^{(k)} \mathcal{L}(\overline{b})$ where $\mathcal{L}(\overline{b}) = 1$ when $\overline{b}$ was initialized from $b_0$.
 
\paragraph{W-POMCP and FS-W-POMCP.}
In both W-POMCP and FS-W-POMCP, we represent the current root-node belief estimate with an offline weighted filter $\overline{b}$ that we update independently of the search tree instead of using the unweighted online particles $B$ stored inside the tree. We provide the pseudo-code for the SIR filter that updates $\overline{b}$ in 
\ifappendix
    \cref{app:wpf}.
\else
    \app~F.2.
\fi
Additionally, FS-W-POMCP maintains statistics in each action node for the actions of pairs of agents instead of all joint actions, as explained previously for FS-POMCP. 

\paragraph{Particle filtering in MPOMDPs.}
In MPOMDPs, the observation signal becomes increasingly sparse as the number of agents increases, as it commonly depends on the probability of all individual observations. 
This can result in an impoverishment of the particles.
Comparably, the likelihood 
of matching the received joint observation in the rejection update is small for unweighted filters in larger observation spaces.
If the particle filter reaches a deprived state where no particles remain, the planner defaults to a baseline policy.

\subsection{Particle Filtering in a Coordination Graph}\label{sec:scalable_pf_method}
To increase the scalability and decrease the chance of deprivation of the particle filter in large observation spaces,
we introduce a general filtering approach for $\overline{b}$ based on the structure of a coordination graph $(\mathcal{V}, \mathcal{E})$, independent of the online planning algorithm.
This method applies to both FT-POMCP~(\cref{sec:fv_pomcp}) as well as FT-PFT (introduced in \cref{sec:factored_pft}).
We exploit the structure in the following way.
For every edge $e \in \mathcal{E}$, we introduce a separate particle filter $\overline{b}_e$ with $K_e$ particles.
We choose $K_e$ such that $K = \sum_e K_e$.
This method makes the following assumption.
\begin{assumption}\label{as:individual_obs} Individual observations probabilities, as given by the individual observation model~$\mathcal{O}_i\colon\mathcal{S} \times\mathcal{A}\rightarrow\Delta(\Omega_i)$, are conditionally independent given the successor state and the previous action.
Therefore, we write the observation model as the product of individual observation probabilities: %O(o_i \given s', \joint{a})$, with $o_i \in \Omega_i$ and 
$
    \mathcal{O}(\joint{o} \given s',\joint{a})
    = \prod_{i\in\mathcal{I}}\mathcal{O}_i(o_i \given s', \joint{a}),
$
with $o_i\in\Omega_i$.
\end{assumption}
\noindent
Note that we condition the individual observations on the joint state and action instead of the assumption of \textit{observational independence} of ND-POMDPs~\citep{DBLP:conf/ijcai/NairVTY05} and,
distinctly from \textit{factored beliefs} \citep{DBLP:conf/nips/MessiasSL11} and \textit{factored particle filtering} \citep{DBLP:conf/uai/NgPP02},
we do not assume any state space factorization.

\paragraph{Local updates.}
Using \cref{as:individual_obs}, we update the particle filters for the edges 
by the local part of the observation space $\joint{o}_{e} \in \Omega_e$.
For an edge $e = (i, j)$, we retrieve the local observation $\joint{o}_e$ by taking the individual observations $o_i, o_j$ from the joint observation $\joint{o}$.
Then, we change the importance weights of the filtering procedure to be based on the \emph{local observation probability} $\mathcal{O}_e(\joint{o}_e \given s', \joint{a})$, instead of the joint observation probability $\mathcal{O}(\joint{o} \given s', \joint{a})$. For each edge $e$, the local observation probability $\mathcal{O}_e(\joint{o}_e \given s', \joint{a}) = \prod_{i\in e} \mathcal{O}_i(o_{i} \given s', \joint{a})$ is the probability of observing $o_i \in \Omega_i$ for each agent $i$ in $e$. 
Consider an offline \emph{weighted} filter $\overline{b}_e=\{(s^{(k)}, w_e^{(k)})\}_{k=1}^{K_e}$ and joint observation $\joint{o}$. We first propagate the particle ${s'}^{(k)} \sim \mathcal{T}(\cdot \given s^{(k)}, \joint{a})$ and then compute the new importance sampling weight ${w_e'}^{(k)} \propto {w_e}^{(k)} \mathcal{O}_e(\joint{o}_e \given s', \joint{a})$, where $\joint{o}_e$ is the local part of the observation $\joint{o}$. Every $\overline{b}_e$ approximates the belief for each $s\in\mathcal{S}$ as:
$
\overline{b}_e(s) = \sum_{k=1}^{K_e} \nicefrac{w^{(k)}_e \delta(s, s^{(k)})}{\sum_{l=1}^{K_e} w_e^{(l)}},
$
where $\delta$ is the Kronecker delta function.
Now, $\overline{b}_e$ represents the particle-belief representing the history $\joint{h}^{e}$ of joint actions $\joint{a}$ and local observations $\joint{o}_e$ for edge $e$, and $\mathcal{L}(\overline{b}_e) = 
\sum_{k=1}^{K_e} w_e^{(k)}$ its likelihood.
We provide more details in 
\ifappendix
    \cref{app:wpf}.
\else
    \app~F.2.
\fi
\paragraph{Ensembling.}
Since we run multiple particle filters in parallel, we must decide how to fuse these beliefs together for sampling.
We propose to treat this set of beliefs as an ensemble. We determine a procedure for sampling the ensemble for every simulation iteration.
We use the likelihood of the weighted particle filter update as a statistic for the quality of the belief approximation \citep{DBLP:conf/atal/KattOA19}. Intuitively, we sample more often from higher-quality filters. More precisely, we give filters that contain particles with a higher probability of generating the true observation a higher chance of getting sampled.
We sample from the set of filters
with probabilities proportional to the likelihoods of the particle filters $\mathcal{L}(\overline{b}_e)$ of the edges: 
$
    s \sim \overline{b}_e \text{ w.p. } \nicefrac{\mathcal{L} (\overline{b}_e)}{\sum_{e'\in\mathcal{E}} \mathcal{L} (\overline{b}_{e'})}.
$
Altogether, this ensemble particle-filter results in the following approximation $\overline{b}$ of the belief $b$ for each $s\in\mathcal{S}$:
$
b(s) \approx \overline{b}(s) = \sum_{e\in\mathcal{E}} \nicefrac{\mathcal{L}(b_e)}{\sum_{e'} \mathcal{L}(b_{e'})} \overline{b}_e(s).
$
FT-W-POMCP and FT-PFT maintain an offline belief estimate $\overline{b}$ consisting of the offline local beliefs $\overline{b}_e$ for each $e$.

\paragraph{Limitation.} 
Our proposal distribution remains $\mathcal{T}$ across the ensemble, but our observation distributions are the local function $\mathcal{O}_e$ for every filter $\overline{b}_e$. Therefore, these local filters will be biased towards the posterior $p(s\given \joint{h}^e, b_0)$ instead of $p(s\given\joint{h}, b_0)$, where, as before, $\joint{h}^{e}$
is the history of factored observations $\joint{o}_e$ and joint actions $\joint{a}$.
Since each filter considers only local observations, the local filters cannot recover a joint belief that depends on all agents~\citep{DBLP:journals/ras/CapitanMCO11}.

%%%%%%%%%%%%%%%%%%%%%%%%%%%
% Particle filtering done %
%%%%%%%%%%%%%%%%%%%%%%%%%%%

\section{Sparse-PFT with Value Factorization}\label{sec:factored_pft}
In this section, we lift Sparse-PFT
to MPOMDPs.
We propose extensions that exploit the factorization of the action space as 
in \cref{sec:fv_pomcp}.
Firstly, we introduce a particle belief approximation and Sparse-PFT for MPOMDPs. 
Then, we introduce variants with \emph{factored statistics} (\textbf{FS-PFT}) and \emph{factored trees} (\textbf{FT-PFT}) to combat large action spaces.
\paragraph{Particle approximation.}
For POMDPs, it is natural to consider a fully observable belief-MDP, whose state space is the beliefs, and the action space is unchanged~\citep{DBLP:conf/aaai/CassandraKL94}. 
The same construction for an MPOMDP yields a belief-MMDP. 
Particle-belief-MDPs approximate belief-MDPs~\citep{DBLP:conf/ijcai/LimTS20,lim2023optimality}.
Similarly, we introduce the particle-belief-MMDP as an approximation of an MPOMDP:
\begin{definition}[\textbf{PB-MMDP}]\label{def:pbmmdp}
    The \emph{Particle-Belief-MMDP} for an MPOMDP $\mathcal{M} = \tuple{\mathcal{I}, \mathcal{S}, \mathcal{A}, \mathcal{T}, r, \Omega, \mathcal{O}, \gamma}$ is a tuple $\mathcal{M}' = \tuple{\mathcal{I}, \Sigma, \mathcal{A}, \tau, \rho, \gamma}$ with
    \emph{states} $\Sigma =  (\mathcal{S} \times \mathbb{R}^+)^C$ consisting of online weighted particle beliefs $\tilde{b} = \{(s^{(k)}, w^{(k)})\}_{k=1}^C$ encoded by $C$ particles,
    the \emph{transition density function} $\tau\colon \Sigma \times \mathcal{A} \rightarrow \Delta(\Sigma)$ 
    defined by $\tau(\tilde{b}' \given \tilde{b}, \joint{a}) = \sum_{\joint{o}\in\Omega} \Pr(\tilde{b}'\given \tilde{b}, \joint{a}, \joint{o})\Pr(\joint{o}\given \tilde{b}, \joint{a})$,
    and the \emph{reward function} $\rho \colon \Sigma \times \mathcal{A} \rightarrow \mathbb{R}$ defined by $\rho(\tilde{b}, \joint{a}) = \nicefrac{\sum_{k} w^{(k)} \mathcal{R}(s^{(k)}, \joint{a})}{\sum_{l} w^{(l)}}$. 
\end{definition}
\noindent
Simulating the PB-MMDP requires us to update the associated generative model. We simulate particle beliefs $\tilde{b}$ of size $C$ instead of individual states to estimate $Q$.
Consequentially, the generative model $\mathcal{G}_{PF} \colon \Sigma \times \mathcal{A} \rightarrow \Sigma \times \mathbb{R}$ updates the state based on the action and returns the particle-based reward $\rho$ as specified above. 
This extension increases the complexity of the generative model by a factor $\mathcal{O}(C)$.
\subsection{Sparse Particle Filter Tree}
Sparse-PFT is an application of UCT to the PB-MMDP. 
While it was designed for continuous state spaces, the fact that the tree branches on a fixed number of belief nodes instead of the number of joint observations is beneficial in our setting.
Sparse-PFT constructs a sparse particle-belief tree incrementally during a forward search by allowing each action node to expand up to $C$ particle-belief nodes. 
The particle-belief nodes correspond to the states of the particle-belief MMDP (\cref{def:pbmmdp}).
The root particle-belief $\tilde{b} \gets \left\{(s^{(k)}, \nicefrac{1}{C})\right\}_{k=1}^C \sim \overline{b}$ is sampled at every simulation iteration from the current offline belief $\overline{b}$.
Following our separation of online and offline beliefs, the number of particles in the offline belief $|\overline{b}| \gg C$ can be much greater than the simulated belief inside the tree $\tilde{b}$.
If the number of children $|\textrm{Ch}(\tilde{b}, \joint{a})|$ of the action node is less than $C$, then we simulate the particle-belief through $\mathcal{G}_{PF}$ to obtain the next particle-belief $\tilde{b}'$ and particle-based reward $\rho$.
Otherwise, $\tilde{b}'$ and $\rho$ are sampled uniformly from $\textrm{Ch}(\tilde{b}, \joint{a})$.
We continue the simulation and traverse the particle-belief tree until we reach a leaf node or a predetermined maximum depth. If we reach a leaf node, a rollout is performed.
Scalability is partially addressed because the branching factor of the belief nodes is independent of the observation size. However, a full enumeration of the action space is still required for selecting actions according to UCB1, which is impractical in MPOMDPs.
\subsection{Sparse-PFT for MPOMDPs}\label{sec:fsandftpft}
We introduce two extensions to improve upon the weakness of Sparse-PFT when operating with large action spaces.
\paragraph{Factored statistics (FS-PFT).}\label{par:fs-pft}
We propose to keep factored action statistics in the nodes of the particle filter tree, similar to FS-POMCP.
In addition to the node visit count $N(\tilde{b})$, we maintain sets of statistics $Q_e(\tilde{b}, \joint{a}_e)$, $N(\tilde{b}, \joint{a}_e)$ in every particle filter belief node that predicts the Q-function for every edge $e \in \mathcal{E}$, applying MoE optimization from \cref{eq:moe} directly in the nodes. 
The offline belief $\overline{b}$ is represented by an external weighted particle filter, as in FS-W-POMCP.
Finally, similarly to the previous factored statistics algorithms, a graphical inference algorithm selects the maximal joint action $\joint{a}^\#$ during the search by maximizing over the UCB1 values:
$
\argmax_{\joint{a}^\#} \sum_{e\in\mathcal{E}} \ucb(Q_e(\tilde{b}, \joint{a}_e), N(\tilde{b}), N(\tilde{b}, \joint{a}_e)).
$
\paragraph{Factored trees (FT-PFT).}\label{par:ft-pft}
Additionally, we introduce the construction of multiple particle-belief trees in parallel, one for each $e \in \mathcal{E}$. %, which we call FT-PFT. 
These trees have a local action space $\joint{a}_e$ and maintain statistics $Q_e(\tilde{b}, \joint{a}_e)$, $N_e(\tilde{b})$, $N_e(\tilde{b}, \joint{a}_e)$ for the agents associated with the edge. 
Since particle filter trees do not explicitly branch on observations, only the action space is factored inside the trees. We use a single joint particle belief step in each layer to reduce overhead. 
Thus, every tree is constructed from the same simulated particle filter beliefs.
Although the belief nodes might have the same particles, we maintain independent visit count statistics $N_e$ for each belief node and associated local joint actions $\joint{a}_e\in\mathcal{A}_e$, respectively. 
The inference equation for picking the maximal UCB1 action (using \cref{eq:ucb}) is given by:
$
\argmax_{\joint{a}^\#} \sum_{e\in\mathcal{E}} \ucb(Q_e(\tilde{b}, \joint{a}_e), N_e(\tilde{b}), N_e(\tilde{b}, \joint{a}_e)).
$
The offline belief $\overline{b}$ is maintain identically to FT-W-POMCP (\cref{sec:scalable_pf_method}), by the ensemble of offline beliefs $\overline{b}_e$.
In addition to the above, we suspect the improvement of FT-PFT over Sparse-PFT is an increase in node re-use and search depth due to the smaller factored action space in the trees.

%%%%%%%%%%%%%%%%%%%%%%%%%%%
% Sparse-PFT section done %
%%%%%%%%%%%%%%%%%%%%%%%%%%%

\section{Experimental Evaluation}\label{sec:experiments}
We evaluate the effectiveness of our methods on MPOMDPs with many agents.
Abbreviations follow those in \cref{tab:intro}.
The key question is \textbf{Q1}: \textit{Does the use of coordination graphs (\introterm{CGs}) accelerate online planners for MPOMDPs in general?}
We evaluate this question on three benchmarks, one with a given coordination graph and two with an artificially chosen graph.
Regarding our novel algorithms introduced in \cref{sec:scalable_pf,sec:factored_pft}, we evaluate \textbf{Q2}: \textit{Do (FS/FT)-W-POMCP variants improve over (unweighted) (FS/FT)-POMCP variants}, and, \textbf{Q3}: \textit{Do FS/FT-PFT improve over Sparse-PFT}?

\paragraph{Benchmarks.}
\textsc{FireFightingGraph} \citep[FFG,][]{DBLP:conf/atal/OliehoekSWV08}
has been used to evaluate factored POMCP \citep{DBLP:conf/aaai/AmatoO15}.
Agents stand in a line, and houses are located to the left and right of each agent. 
Agents have two actions: fight fires to their left or right. 
Multi-agent \textsc{RockSample} \citep[MARS,][]{DBLP:journals/ijrr/CaiLHL21} extends single-agent RockSample \citep{DBLP:conf/uai/SmithS04}. 
MARS environments are defined by their size $m$, the number of agents $n$, and the number of rocks $k$, with $k=m=15$.
In \textsc{CaptureTarget}~(CT), agents are tasked with capturing a moving target. We depict results for CT in 
\ifappendix
    \cref{appendix:expeval-plots}.
\else
    \app~A.2.
\fi
Detailed benchmark descriptions are in 
\ifappendix
    \cref{appendix:benchmarks}.
\else
    \app~G.
\fi

\paragraph{Experimental set-up.}
All algorithm variants are implemented in the same Python prototype, published online\footnote{\url{https://zenodo.org/records/10409525}.}.
All code ran on a machine with an Intel(R) Core(TM) i9-10980XE CPU @~3.00GHz and 256~GB~RAM~(8 x 32GB DDR4-3200). 
Our Python wrapper executed episodes in parallel on 34 threads such that each episode had access to $256/34\approx7.5$GB of RAM. 
All reported results are the achieved returns averaged over 100 episodes with error bars representing 95\% confidence intervals. 
We did not run an extensive hyperparameter optimization for any algorithm, and we list the most important parameters in 
\ifappendix
    \cref{tab:hyperparams} of \cref{appendix:expeval-params}.
\else
    Tab. 2 of \app~A.1.
\fi
All algorithms ran with a maximum of 5s and 15s per step on FFG/CT and MARS, respectively. 
If the particle filter belief is deprived at any point in time during the episode, the policy defaults to a random policy. 
We set the number  $K$ of particles in the joint filters such that $K = \sum_e K_e$ in the factored filters, e.g., if we have three edges with $K_e=100$, then the joint counterpart has $K=300$.
For MARS and CT, we chose the CG as a \textit{line} ($n-1$ factors) for odd numbers of agents and a \textit{team} formation ($\nicefrac{n}{2}$ factors) where pairs of agents cooperate for even numbers.
The single-agent algorithms could not run with more than $20$ and $5$ agents on FFG and MARS, respectively.
\begin{figure}[t]
    \centering
    \renewcommand\sffamily{}
    \resizebox{\columnwidth}{!}{\input{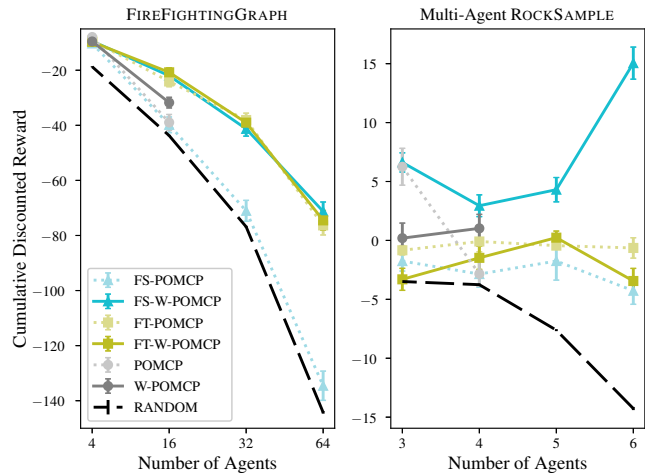}}
    \caption{Performance comparison for POMCP variants with (solid) and without (dotted) weighted particle filtering. 
    }
    \label{fig:pomcpwf}
\end{figure}
\paragraph{Discussion.} Below, we analyze the results as answers to the three questions.
\textbf{Q1.}
We study Q1 across our different set-ups. 
The single-agent algorithms (Sparse-PFT, POMCP, and W-POMCP) are out-scaled by their competitors with value factorization (\cref{fig:pomcpwf,fig:pft}) in FFG and MARS. However, planning on the joint value performs better in settings with fewer agents. 
In MARS and CT, the agents move and thus may coordinate dynamically. Therefore, the desired agent coordination does not induce a \emph{sparse} coordination graph, meaning the CG acts as a heuristic.
The results show that assuming some arbitrary, sparse static graph is helpful, even if this assumes no coordination between agents that, in principle, should coordinate.
We find that the static heuristic performs well when many agents are involved.
Thus, \textbf{CGs (as a heuristic) accelerate planning}.
\noindent
\textbf{Q2.}
FS-W-POMCP outperforms FS-POMCP across all three benchmarks, showing that \textbf{POMCP benefits from offline weighted particle filtering}. The difference between FT-POMCP and FT-W-POMCP is smaller, as FT-POMCP also benefits from the fact that the belief representation, which is offline, consists of local beliefs $\tilde{b}_e$ for each $e$, albeit unweighted (see 
\ifappendix
    \cref{appendix:ft-pomcp}).
\else
    \app~E).
\fi
\noindent
\textbf{Q3.}
Sparse-PFT performs well in MPOMDPs with fewer agents, but with more agents, it runs out of memory or fails to find a good estimate of $Q$ due to its naive enumeration of the action space (\cref{fig:pft}). FS-PFT and FT-PFT do scale to settings with many agents.
Thus, \textbf{CGs alleviate Sparse-PFTs scaling issues in many-agent POMDPs.}
However, they achieve comparable (FFG) or lower (MARS) returns in the settings with few agents.
\noindent
\textbf{Combining CGs with weighted filtering performs well across.}
We cross-evaluate our contributions in 
\ifappendix
    \cref{fig:pftvspomcp} (\cref{appendix:expeval-plots}). 
\else
    (Fig.~3, \app~A.2).
\fi
We find that both factored W-POMCP and PFT algorithm variants are suited for many-agent POMDPs and perform well across FFG and MARS, but POMCP variants generally perform better.
The improvement of FS-W-POMCP over FS-POMCP is consistent. FT-W-POMCP is slightly better on FFG and significantly improves over FT-POMCP in CT, but performs equal or worse on MARS. 
\noindent
\textbf{The action selection method has a noticeable influence on performance.}
In MARS and CT, VE is the best-performing algorithm. However, MP achieves much higher returns in FFG 
\ifappendix
    (\cref{fig:ffg-both}, \cref{appendix:expeval-plots}).
\else
    (Fig.~4, \app~A.2).
\fi
The factored algorithms are sensitive to the method that maximizes over the local predictions, as recently also demonstrated in the fully observable setting~\citep{DBLP:journals/jair/ChoudhuryGMK22}.
\begin{figure}[t]
    \centering
    \renewcommand\sffamily{}
    \resizebox{\columnwidth}{!}{\input{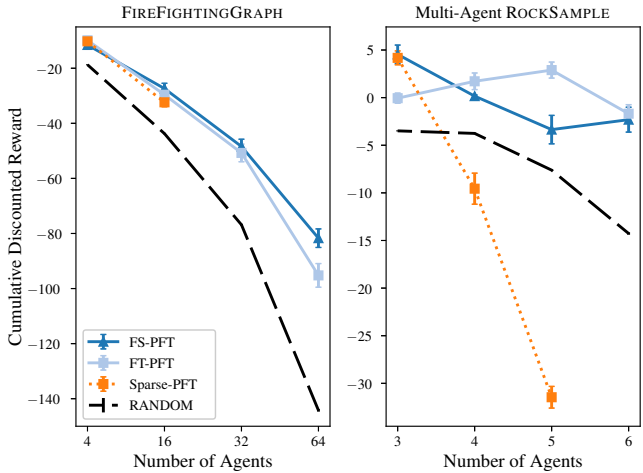}}
    \caption{Comparison between Sparse-PFT (dotted) and our PFT variants with value factorization (solid).}
    \label{fig:pft}
\end{figure}

%%%%%%%%%%%%%%%%%%%%%%%%
% Results section done %
%%%%%%%%%%%%%%%%%%%%%%%%

\section{Related Work}
\paragraph{Multi-agent Markov models.}
MPOMDPs reside in a realm of models for cooperative multi-agent systems with partial observability.
Distributed cooperative systems  \citep[Dec-POMDPs,][]{DBLP:series/sbis/OliehoekA16} remove the communication assumptions of MPOMDPs. However, they are much more computationally complex (doubly exponential), as agents need to reason over each other's policies.
\citet{DBLP:conf/nips/MessiasSL11} considered factored MPOMDPs by assuming shared communication in Dec-POMDPs, computing policies over \textit{factored beliefs}. Additionally, they studied lifting the instantaneous communication assumptions by asynchronous execution~\citep{DBLP:conf/atal/MessiasSL13}.
Our algorithms build on prior work by
\citet{DBLP:conf/aaai/AmatoO15}. Therefore, their work is summarized in \cref{sec:structurepomdps}. 
\citet{DBLP:journals/access/ZhouWZWZ19} introduce a further decentralized MCTS algorithm for transition-independent MPOMDPs. 
\citet{DBLP:journals/jair/ChoudhuryGMK22} consider a fully observable MMDP setting and study action selection under state-dependent coordination graphs. Recently,
MPOMDPs were also studied with barrier functions over the joint belief~\citep{DBLP:conf/cdc/AhmadiSBA19}
and to support multi-object tracking~\citep{DBLP:conf/aaai/NguyenRVR20}.
\paragraph{Single-agent online planning.}
POMCPOW~\citep{DBLP:conf/aips/SunbergK18} and Sparse-PFT~\citep{lim2023optimality}, also in \cref{tab:intro}, are algorithms that improved UCT-based planners for POMDPs with continuous spaces. They, i.a., replaced unweighted belief estimates with importance sampling estimates using weighted particle filters (\cref{sec:scalable_pf}).
We summarize Sparse-PFT in \cref{sec:factored_pft}.
POMCPOW shares characteristics with W-POMCP and Sparse-PFT, simulating single states but maintaining weighted particles in the tree.
DESPOT~\citep{DBLP:journals/jair/YeSHL17} and AdaOPS~\citep{DBLP:conf/nips/WuYZYLLH21} are alternative, orthogonal online planners that are distinguishable from MCTS methods (e.g., POMCP).
DESPOT utilizes a set of deterministic scenarios and heuristic tree searches to reduce variance in the value estimates instead of the independent simulations via MCTS.
Its extensions employ alpha-vectors to fuse similar paths in the tree \citep{DBLP:conf/rss/GargHL19} or (GPU) parallelization in factored simulators \citep{DBLP:journals/ijrr/CaiLHL21}. 
AdaOPS also employs offline and weighted particle filtering. 
Distinctively, it uses adaptive particle filtering \citep{DBLP:conf/nips/Fox01}, which requires a partitioning of the state space into grids. It relies on a full-width search instead of simulations, during which it fuses similar observation branches.
Both algorithms work well with small-sized discrete action spaces. 
However, it is unclear how value factorization from coordination graphs can be incorporated, as both algorithms expand the full action space at each new node instead of picking the most promising action to simulate via the UCB1 policy.
In MPOMDPs, expanding the combinatorial joint action space is impractical.

%%%%%%%%%%%%%%%%%%%%%%%%%%%%%
% Related work section done %
%%%%%%%%%%%%%%%%%%%%%%%%%%%%%

\section{Conclusion}
In this paper, we studied how to simultaneously tackle the belief and value estimation challenges in online planning for MPOMDPs.
We presented extensions of factored POMCP and novel variants of the Sparse-PFT algorithms tailored specifically for many-agent online planning with partial observability.
The empirical evaluation showed the effectiveness of combining weighted particle filtering and value factorization in settings with many agents.
However, it is also clear that planning on the joint value suffices when few agents are involved. 
% Furthermore, in line with \citet{DBLP:journals/aamas/CastelliniOSW21}, we find that arbitrary coordination graphs act as heuristics for scaling value function estimates when there does not exist a natural static structure. 
Future work consists of alleviating the communication assumptions
~\citep{DBLP:conf/aips/SpaanOV08,DBLP:conf/aaai/OliehoekS12,DBLP:conf/atal/MessiasSL13}, exploring extensions for continuous MPOMDPs, or learning the coordination graph~\citep{DBLP:conf/cig/KokHBV05}.

\section*{Acknowledgments}
We would like to thank the anonymous reviewers for their valuable feedback. 
This research has been partially funded by the NWO grant
NWA.1160.18.238 (PrimaVera) and the ERC Starting Grant
101077178 (DEUCE). Additionally, we would like to thank the ELLIS Unit Nijmegen and Radboud~AI for their support.

\bibliography{aaai24}

\ifappendix

    \clearpage
    % \onecolumn
    \appendix
    
    \noindent{\Huge Technical Appendix}\\
% \chapter{Technical Appendix}
This document includes:
\begin{itemize}
    \item \cref{appendix:expeval}: Additional plots and information about hyperparameters of the experimental evaluation.
    \item \cref{sec:pbmmdp-appendix}: Additional details regarding the PB-MMDP approximation (\cref{def:pbmmdp}) of \cref{sec:factored_pft}, from \citet{lim2023optimality}.
    \item \cref{appendix:bias}: Analysis on the approximation induced by value factorization from \citet{DBLP:conf/aaai/AmatoO15}.
    \item \cref{sec:action_selection}: Detailed description of the action selection methods \textit{max-plus} (MP) and \textit{variable elimination} (VE).
    \item \cref{appendix:ft-pomcp}: Details on particle filtering in FT-POMCP.
    \item \cref{sec:pseudocode-appendix}: Pseuodo-codes of the various procedures and algorithms of the main paper.
    \item \cref{appendix:benchmarks}: Detailed benchmark descriptions.
    \item \cref{appendix:trees}: Depictions of the various tree structures.
\end{itemize}

\section{Experimental Evaluation}\label{appendix:expeval}
\subsection{Hyperparameters}\label{appendix:expeval-params}
We list all hyperparameters used across our experimental evaluation in \cref{tab:hyperparams}. The methods for action selection (Act.) are detailed in \cref{sec:action_selection}. Detailed environment (Env) descriptions are given in \cref{appendix:benchmarks}.
\begin{table}[ht]
\centering
\begin{tabular}{@{}llllllll@{}}
 & \multicolumn{6}{l}{\textbf{Hyperparameters}} \\ \cmidrule(l){2-8}
\textbf{Env}     & $s$   & TO & Act. & $c$  & $C$ & $K_e$ & $\gamma$ \\
  \cmidrule(l){1-1}   \cmidrule(l){2-2} \cmidrule(l){3-3} \cmidrule(l){4-4} \cmidrule(l){5-5} \cmidrule(l){6-6} \cmidrule(l){7-7} \cmidrule(l){8-8}
FFG  & 1000  & 5s  & MP & 5    & 10  & 20 & 0.99 \\
% \cmidrule{1-1}\cmidrule(l){2-2} \cmidrule(l){3-3} \cmidrule(l){4-4} \cmidrule(l){5-5} \cmidrule(l){6-6} \cmidrule(l){7-7} \cmidrule(l){8-8}
MARS & 10000 & 15s & VE & 1.25 & 10  & 100 &  0.95 \\
% \cmidrule{1-1}\cmidrule(l){2-2} \cmidrule(l){3-3} \cmidrule(l){4-4} \cmidrule(l){5-5} \cmidrule(l){6-6} \cmidrule(l){7-7} \cmidrule(l){8-8}
CT & 10000 & 5s & VE & 0.5 & 10 & 100 & 0.95
\end{tabular}
\caption{Hyperparameters across runs. The computational budget per search is constrained by reaching either $s$, the maximum number of simulations, or the timeout TO. \emph{Max-Plus} (MP) performed best for FFG and \emph{Variable Elimination} (VE) for MARS and CT. $c$ is the UCB1 exploration constant (\cref{eq:ucb}), and $C$ is the number of particles simulated inside PFT variants (\cref{sec:factored_pft}). Finally, $K_e$ is the number of particles per edge $e$ (\cref{sec:scalable_pf_method}, note that $K=\sum_e K_e$) and $\gamma$ is the discount factor.}
\label{tab:hyperparams}
\end{table}

\subsection{Additional plots}\label{appendix:expeval-plots}
We evaluate across both our new methods on both FFG and MARS in \cref{fig:pftvspomcp}. FS-W-POMCP is the best-performing algorithm out of these two benchmarks. FT-W-POMCP, FS-PFT, and FT-PFT perform similarly on MARS. POMCP variants perform better on FFG.

\begin{figure}[ht]
    \centering
    \renewcommand\sffamily{}
    \resizebox{\columnwidth}{!}{\input{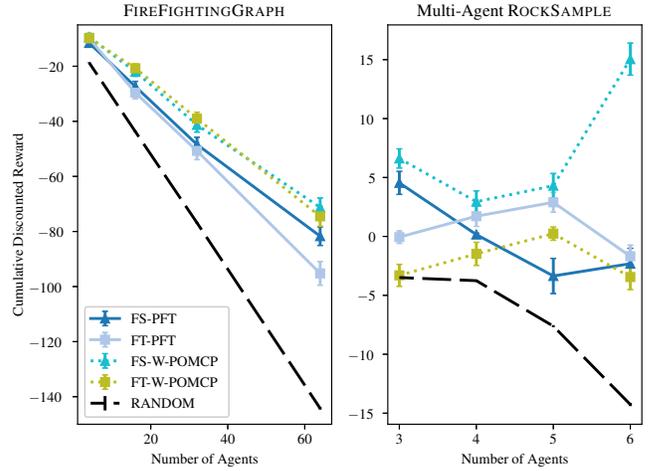}}
    \caption{Performance comparison for our FS/FT-W-POMCP (dotted) and FS/FT-PFT (solid) methods.}
    \label{fig:pftvspomcp}
\end{figure}

\paragraph{Firefighting.}
We plot the comparison of multiple algorithms across the action selection methods in \cref{fig:ffg-both} over FFG. We observe that Max-Plus performs much better in this setting. Interestingly enough, Variable Elimination outperforms Max-Plus on MARS. This evidences that the action selection procedure affects the performance of planning algorithms when employing value factorization over coordination graphs.

\begin{figure}[ht]
    \centering
    \renewcommand\sffamily{}
    \resizebox{0.45\textwidth}{!}{\input{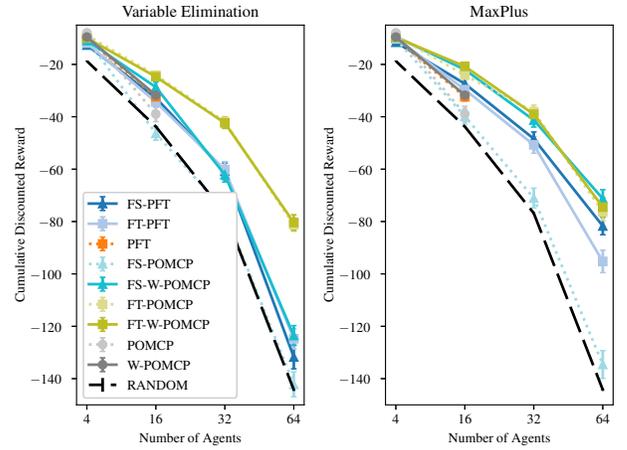}}
    \caption{Comparison across algorithm variants on \textsc{FireFightingGraph} for both Max-Plus and Variable Elimination.}
    \label{fig:ffg-both}
\end{figure}

In \cref{fig:ffg-over-sims}, we plot the performance of the best-performing algorithms using Max-Plus on FFG across the maximum number of permitted simulations. We see that factored algorithms scale much faster, but all algorithms can find good returns for the setting with few agents. Furthermore, only factored algorithms achieve high returns in the setting with many agents. Only FT-POMCP performs competitively with our algorithms but is marginally outperformed by FS-W-POMCP on the 64-agent setting. FS-W-POMCP, FT-W-POMCP, FT-PFT, and FT-POMCP perform best on FFG on the various agent settings and across different numbers of maximal simulation budgets. 

\begin{figure}[ht]
    \centering
    \renewcommand\sffamily{}
    \resizebox{0.45\textwidth}{!}{\input{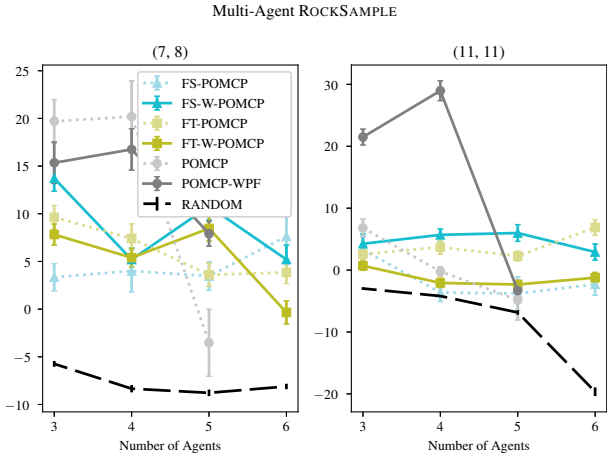}}
    \caption{Performance on two smaller maps of Multi-Agent \textsc{RockSample}, comparing W-POMCP variants to POMCP variants as in \cref{fig:pomcpwf} of the main paper.}
    \label{fig:mars-pomcp-other-maps}
\end{figure}

\paragraph{Rock sampling.}
We ran the MARS experiment on two other maps, namely with $(m=7, k=8)$ and $(m=11,k=11)$. We plot the results in additional versions of the plots of the main paper. Namely, \cref{fig:mars-pomcp-other-maps} extends \cref{fig:pomcpwf}, \cref{fig:mars-pft-other-maps} extends \cref{fig:pft}, and \cref{fig:mars-pftvspomcp-other-maps} extends \cref{fig:pftvspomcp}. 
The flat algorithm variants perform much better on the smaller map $(m=7, k=8)$ due to the decreased action space size. 
On this map, the flat variants are actually able to run the 6-agent version. 
We observe fluctuations in performance across the factored algorithms when transferring from even to odd numbers of agents, since even (\textit{team}) and odd (\textit{line}) are given different coordination graphs. 
This indicates that some coordination graphs perform better than others.

\begin{figure}[ht]
    \centering
    \renewcommand\sffamily{}
    \resizebox{0.45\textwidth}{!}{\input{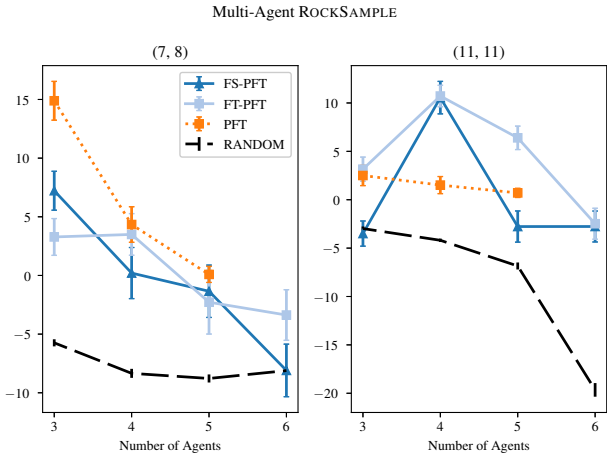}}
    \caption{Performance on two smaller maps of Multi-Agent \textsc{RockSample}, comparing Sparse-PFT (dotted) to FS/FT-PFT (solid) as in \cref{fig:pft} of the main paper.}
    \label{fig:mars-pft-other-maps}
\end{figure}

\begin{figure}[ht]
    \centering
    \renewcommand\sffamily{}
    \resizebox{0.45\textwidth}{!}{\input{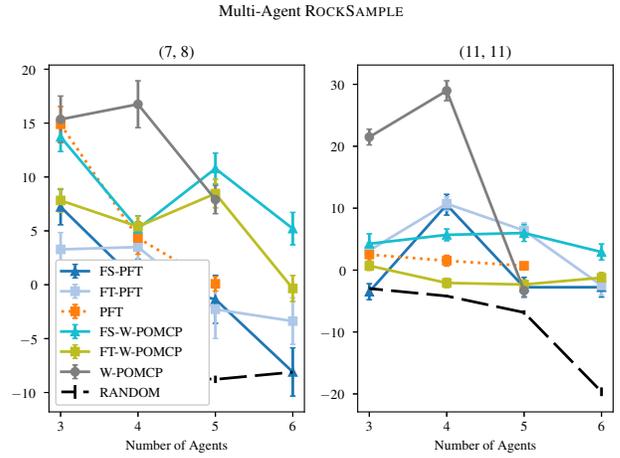}}
    \caption{Performance on two smaller maps of Multi-Agent \textsc{RockSample}, comparing all our weighted FS/FT variants as in \cref{fig:pftvspomcp}.}
    \label{fig:mars-pftvspomcp-other-maps}
\end{figure}

\paragraph{Capturing a target.}
We provide a similar experiment for POMCP variants in \cref{fig:ct-pomcp}. We observe that both FT-W-POMCP and FS-W-POMCP outperform FT-POMCP and FS-POMCP, respectively. We can clearly observe on this benchmark that weighted particle filtering is essential for good performance. Even W-POMCP performs relatively well, which can be explained by the fact that the action space of CT is manageable, and thus, value factorization is not as essential. We do see that FT-W-POMCP and FS-W-POMCP outperform W-POMCP, except for the setting with 4 agents, where it beats FS-W-POMCP by a slight margin. 
All PFT variants perform poorly on CT. The PB-MMDP approximation incurs an $\mathcal{O}(C)$ increase in the computational cost of a simulation step call. The CT benchmark (1) has a relatively expensive simulation step which amplifies the increased computational cost, and (2) requires a significant search depth due to reward sparsity, which explains the poor performance of Sparse-PFT variants. A similar observation was made by \citet{DBLP:conf/aips/SunbergK18}.

\begin{figure}[htbp]
    \centering
    \renewcommand\sffamily{}
    \resizebox{0.45\textwidth}{!}{\input{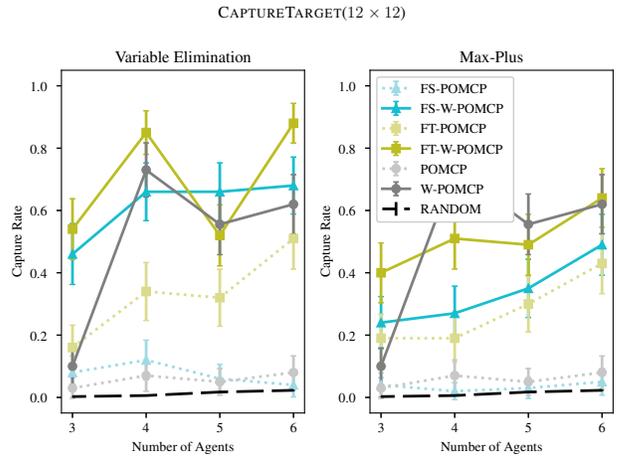}}
    \caption{Performance across two action selection methods VE and MP (\cref{sec:action_selection}) on \textsc{CaptureTarget} of POMCP variants without (dotted) and with weighted particle filtering (solid).}
    \label{fig:ct-pomcp}
\end{figure}

\begin{figure}[htbp]
    \centering
    \renewcommand\sffamily{}
    \resizebox{0.45\textwidth}{!}{\input{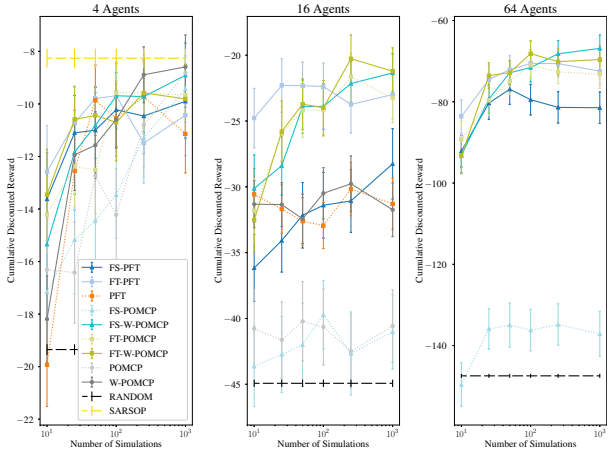}}
    \caption{Performance across the number of permitted simulations on \textsc{FireFightingGraph}, only showing the best-performing algorithms. Our variants are depicted with solid lines, and existing algorithms with dotted lines. As a proxy for an optimum, we ran SARSOP \citep{DBLP:conf/rss/KurniawatiHL08} offline for 30 minutes on the 4-agent setting before running the same empirical evaluation as the online planners.}
    \label{fig:ffg-over-sims}
\end{figure}

\section{Particle Belief-based approximations}\label{sec:pbmmdp-appendix}
In this section, we write out the full definitions of the PB-MMDP model introduced in \cref{def:pbmmdp} of the main paper.

\begin{definition}[\textbf{PB-MMDP}]\label{def:pbmmdp_appendix}
    The particle-belief-MMDP corresponding to an MPOMDP $\mathcal{M} = \tuple{\mathcal{I}, \mathcal{S}, \mathcal{A}, \mathcal{T}, \mathcal{R}, \Omega, \mathcal{O}, \gamma}$ as in \cref{def:mpomdp} is a tuple
    $\tuple{\mathcal{I}, \Sigma, \mathcal{A}, \tau, \rho, \gamma}$, with:
    \begin{itemize}
    \item $\mathcal{I}$, $\mathcal{A}$, $\gamma$ as in the original MPOMDP $\mathcal{M}$ (\cref{def:mpomdp});
    \item $\Sigma$, the state space over particle beliefs $\tilde{b} = \{(s^{(k)}, w^{(k)})\}_{k=1}^C$, where $s^{(k)} \in \mathcal{S}$, $w^{(k)} \in \mathbb{R}^+$ is the associated weight of state particle $s^{(k)}$, and $C = |\tilde{b}|$ the number of particles;
    \item $\tau$, the transition density function $\tau(\tilde{b}' \given \tilde{b}, \joint{a}) \colon \Sigma \times \mathcal{A} \to \Delta(\Sigma)$.
    Given that the importance weights are updated according to the observation density:
    \begin{equation*}
        {w'}^{(k)} \propto {w}^{(k)} \mathcal{O}(\joint{o} \given {s'}^{(k)},\joint{a}),
    \end{equation*}
    Then, transition density $\tau$ is defined as:
    \begin{equation}\label{eq:tau}
        \tau(\tilde{b}' \given \tilde{b}, \joint{a}) \equiv \sum_{\joint{o}\in\Omega} \Pr(\tilde{b}'\given \tilde{b}, \joint{a}, \joint{o})\Pr(\joint{o}\given b, \joint{a})
    \end{equation}
    \item $\rho$, the reward function $\rho(\tilde{b}, \joint{a}) \colon \Sigma \times \mathcal{A} \to \mathbb{R}$ defined by $\rho(\tilde{b}, \joint{a}) = \frac{\sum_{k} w^{(k)} \mathcal{R}(s^{(k)}, \joint{a})}{\sum_{l} w^{(l)}}$, with $\mathcal{R}$ the reward function of the MPOMDP $\mathcal{M}$.
    If the original reward function $\mathcal{R}$ is bounded by $\mathcal{R} \leq \mathcal{R}_{max}$ then $\rho$ is bounded by $||\rho||_\infty \leq \mathcal{R}_{max}$ as the importance weights satisfy $\sum_k w^{(k)} = 1$.
    \end{itemize}
\end{definition}
\paragraph{Defining the transition density.}
The first term $\Pr(\tilde{b}'\given \tilde{b}, \joint{a}, \joint{o})$ in \cref{eq:tau} is the conditional transition density given observation $\joint{o}$. This probability is synonymous with the belief update function $\upsilon$ computed with Bayes' theorem \citep{DBLP:conf/aaai/CassandraKL94}, albeit for a particle belief. The state transition updates for each particle in the particle belief are independent of each other, and the update of the likelihood weight is deterministic given $s^{(k)}, {s'}^{(k)}, \joint{a}, \joint{o}$. Furthermore, it is only non-zero when $\tilde{b}' = \{({s'}^{(k)}, {w'}^{(k)})\})_{k=1}^C$, such that the calculation simplifies to the product of the individual transition densities given that the likelihood update matches \citep{lim2023optimality}:
\begin{equation*}\textstyle
\Pr(\tilde{b}'\given \tilde{b}, \joint{a}, \joint{o}) = 
    \begin{cases}
      \prod_{k=1}^C\mathcal{T}({s'}^{(k)}\given {s}^{(k)}, \joint{a}) & \text{if}\ P({w'}^{(k)}), \text{ and,} \\
      0 & \text{otherwise,}
    \end{cases}
\end{equation*}
Where $P({w'}^{(k)}) = \forall_k \colon {w'}^{(k)} = w^{(k)}\mathcal{O}(\joint{o}\given {s'}^{(k)}, \joint{a})$.
The second term $\Pr(\joint{o}\given \tilde{b}, \joint{a})$ is the observation likelihood given a particle belief and a joint action. 
We need to define this probability for a particle belief.
The likelihood is equivalent to the weighted sum of the observation likelihoods conditioned on the probability that this observation was generated from state particle $s^{(k)}$:
\begin{equation*}\textstyle
    \Pr(\joint{o}\given \tilde{b}, \joint{a}) = \cfrac{\sum_{k=1}^C w^{(k)}\left(\sum_{s'}\mathcal{O}(\joint{o}\given {s'}, \joint{a})\mathcal{T}(s'\given s^{(k)}, \joint{a}) \right)}{\sum_{l=1}^C w^{(l)}}.
\end{equation*}
Although this transition density is difficult to calculate exactly, it is possible to sample from it. Given that we do not require transition probabilities of the PB-MMDP but mere simulation access, we can perform UCT on this model.
\paragraph{Approximation quality.} 
\citet{lim2023optimality} show that, under a set of assumptions, the approximation error of the optimal value of a PB-MDP to a POMDP is small with high probability. In addition to a finite horizon $H$, and a finite action space $\mathcal{A}$, the proof relies on a reward function $\mathcal{R}$ of the POMDP (multi-agent in our case) that is bounded and (in the continuous case) Borel measurable, such that $||\mathcal{R}||_\infty \leq \mathcal{R}_{\max} < +\infty$, and the maximum value satisfies $V_{\max} \equiv \cfrac{\mathcal{R}_{\max}}{1-\gamma}$, with $V_{\max} < +\infty$. Finally, their proof assumes the state and observation spaces are continuous, but note that it also holds for discrete spaces by replacing the integrals with (Riemann) sums, as long as the infinite R\'enyi divergence $d_\infty(\mathcal{P}^d\,||\,\mathcal{Q}^d) \leq d^{max}_\infty < +\infty$ between the target $\mathcal{P}^d$ and the proposal $\mathcal{Q}^d$ distributions is bounded above almost surely by $d^{\max}_\infty$, for depth $d$ up to a given maximum depth $D$, with $0 \leq d \leq D$.
$\mathcal{P}^d$ is the normalized measure incorporating the probability of observing the past $d$ observations given the past $d$ states and $d-1$ actions, and $\mathcal{Q}^d$ is such measure on the past states conditioned on the past actions.
More intuitively, this means that the ratio between the marginal and conditional observation probability cannot become infinitely large.
For more details  
see \citet{lim2023optimality}.

\section{Convergence}\label{appendix:bias}
\subsection{Mixture of Experts}
\citet{DBLP:conf/aaai/AmatoO15} show that the MoE optimization introduces a policy-dependent bias term $B_\pi(\joint{a})$ given sample policy $\pi$. This bias is introduced by the overlap of the edges, where agent $i$ might be part of both edge $e$ and $e'$. 
Since their analysis is \textit{stateless} and focused on the predictions of the Q-value, it encompasses both FS-POMCP and the new variants FS-W-POMCP and FS-PFT.
We denote the neighboring edges that overlap in the agents of $e$ as $\Gamma(e) = \{e' \in \mathcal{E} \setminus \{e\} \mid e \cap e' \not= \emptyset\}$, where $e \cap e'$ is non-empty if there exists an agent $i$ with $i \in e$ and $i \in e'$.
Furthermore, we repeat that we assume the experts are weighted uniformly, \ie, $\omega_e = \nicefrac{1}{|\mathcal{E}|}$ such that these weights can be omitted. We summarize the analysis by \citet{DBLP:conf/aaai/AmatoO15} here.

The MoE estimate approaches the true value plus the bias term $\hat{Q}(\joint{a}) \approx Q(\joint{a}) + B_\pi(\joint{a})$. Let $\joint{a}_{e'\setminus e}$, specified by $\joint{a}$, be the actions of the agents in $e'$ that are not in $e$, and, conversely, let $\joint{a}_{e'\cap e}$ be the actions of the agents that participate in both $e$ and $e'$. Then, the bias term is defined as:
\begin{equation*}
    B_\pi(\joint{a}) \triangleq \sum_{e}\sum_{e'\not=e}\sum_{\joint{a}_{e'\setminus e}} \pi(\joint{a}_{e'\setminus e}\given \joint{a}_e) Q_{e'}(\joint{a}_{e'\setminus e}, \joint{a}_{e'\cap e}).
\end{equation*}
Bias alone is not an issue, but differing biases per joint action can result in non-optimal action selection. Fortunately, their proof also encompasses that the bias terms between actions are bounded by $\forall_{\joint{a},\joint{a}'} \colon |B_\pi(\joint{a}) - B_\pi(\joint{a}')| \leq\epsilon$ if the Q-function is sufficiently structured. If value functions do not overlap, MoE recovers the maximal joint action. If they do overlap, then we have the following.
\begin{theorem}\label{th:factoredQ}
    If for edges with overlap $e$, $e'$, and any two  $\joint{a}_{e'\cap e}, \joint{a}'_{e'\cap e} \in \mathcal{A}_{e'\setminus e}$, with $\mathcal{A}_{e'\setminus e}$ the set of actions with overlap, the true value function $Q$ satisfies:
    \begin{align*}
        \forall_{\joint{a}_{e'\setminus e}} &\colon Q_{e'}(\joint{a}_{e'\setminus e}, \joint{a}_{e'\cap e}) - Q_{e'}(\joint{a}_{e'\setminus e}, \joint{a}'_{e'\cap e}) \\ &\leq \cfrac{\epsilon}{|\mathcal{E}|\cdot |\Gamma(e)|\cdot |\mathcal{A}_{e'\setminus e}|\cdot \pi(\joint{a}_{e'\setminus e})},
    \end{align*}
    then MoE optimization will return an $\epsilon$-optimal joint action in the limit. 
\end{theorem}
\begin{proof}
    See the proof for Theorem 6 in the appendix of \citet{DBLP:conf/aaai/AmatoO15}.
\end{proof}
\subsection{Factored Trees}
\paragraph{FT-POMCP.}
\citet{DBLP:conf/aaai/AmatoO15} show that the FT-POMCP variant might diverge when UCB1 with exploration constant $c=0$ is employed. In that case, the algorithm corresponds with a class of Monte Carlo control, such as SARSA(1), with linear regression. 
These settings divert in general \citep{DBLP:conf/ijcnn/FairbankA12b}.
This can be explained by the fact that instead of the full history $\joint{h}$, the local history $\joint{h}_e$ is not Markov if the factors are not independent. During the search, the exploration constant $c$ of the UCB1 policy is not zero, but it is zero when selecting the final action greedily to execute in the environment. Although UCB1 is a greedy policy with respect to the upper confidence bounds of the Q-values, it is not clear how the exploration bonus of the UCB1 policy affects the convergence of the algorithm. Additionally, UCB1 is not a policy for the local history as it is defined for the full action space. However, we follow \citet{DBLP:conf/aaai/AmatoO15} in stating that the factored trees variants might produce high-quality results in practice, especially when the problem benefits from an increased search depth. Such search depth is made possible due to the additional generalization offered by considering local histories.

\paragraph{FT-PFT.} For FT-PFT, we do not have such a strong result. In this algorithm, an estimate of the joint belief is propagated in each tree, but these trees only branch on the factored action space $\mathcal{A}_e$, where $e = (i, j)$. We can, however, make a similar observation on the UCB1 policy, as it is a policy for the full action space, which we calculate based on statistics from the local tree for $e\in\mathcal{E}$, ranging over $\mathcal{A}_e$. Thus, UCB1 is now also employed edge-wise instead of over the joint action space.

\section{Action Selection}\label{sec:action_selection}
For our MCTS-based approaches, we require the UCB1 action (see \cref{eq:ucb} in \cref{sec:mcts-pomcp} for details) at every simulation step during the forward search. Additionally, we need to pick the final action after searching. Due to the induced factorization of the CG, we need to maximize over local combinations. In this section, we summarize two graphical inference methods, \emph{variable elimination} and \emph{max-plus}, that can solve the maximization problem introduced by the MoE optimization (\cref{eq:max-over-moe}).
Examples of this problem include the inference equations of the FS and FT variants of POMCP and PFT in the main paper.

\paragraph{Exact.}
Variable Elimination (VE) is an exact algorithm for inference in probabilistic graphical models.
The global best action is computed by eliminating agents from the graph and adding the resulting conditional payoff functions to the graph.
VE has a run-time complexity of $\mathcal{O}(|\mathcal{V}|\cdot|\mathcal{A}_{\max}|^w)$ \citep{DBLP:conf/aaai/AmatoO15}, where $|\mathcal{V}|$ is the number of agents, $|\mathcal{A}_{\max}|$ is the cardinality of the largest action-set, and $w$ is the induced width of the CG \citep{DBLP:series/synthesis/2007Vlassis}. 
The induced width is synonymous with the size of the largest clique to be computed during node elimination. VE is very efficient at computing maximal actions but suffers from an exponential worst-case complexity in dense graphs with many agents. The agent elimination ordering has a large effect on the computational complexity. In our experiments, we use the ordering induced by sorting the nodes by degree.

Following \cite{DBLP:conf/nips/GuestrinKP01}, the algorithm maintains a set $\mathcal{F}$ of functions, with $\mathcal{F} = \{Q_{e_1}, Q_{e_2}, \ldots, Q_{e_{|\mathcal{E}|}}\}$ initially, where $Q_e$ is the value-function of the edge indicated by $e \in \mathcal{E}$ in the graph, with $e=(i,j)$. The algorithm then proceeds as follows:
\begin{enumerate}
    \item Pick any non-eliminated agent $i \in \mathcal{V} \setminus \mathcal{X}$ ($\mathcal{X} = \emptyset$ initially). If $\mathcal{V} \setminus \mathcal{X} = \emptyset$, then terminate.
    \item Aggregate all $Q_e \in \mathcal{F}$ for which $\mathcal{A}_i \in \text{Scope}[Q_e]$ into $\mathcal{H}$, i.e., the $Q_e$'s that range over the action space $\mathcal{A}_i$ of agent $i$.
    \item Add function $f = \max_{a_i\in\mathcal{A}_i}\sum_{g\in\mathcal{H}}g$ to $\mathcal{F}$, with $\text{Scope}[f] = \bigcup_{g'\in\mathcal{H}} \text{Scope}[g'] \setminus \{\mathcal{A}_i\}$.
    \item Add agent $i$ to the set of eliminated agents $\mathcal{X} \gets \mathcal{X} \cup \{i\}$, and repeat from 1.
\end{enumerate}
A reverse pass, i.e., with the reverse of the order used by the procedure above, decides the optimal action for each agent by maximizing over the newly introduced functions in $\mathcal{F}$.

\paragraph{Anytime.}
The Max-Plus (MP) algorithm for computing the maximum a posteriori (MAP) configuration is analogous to belief propagation in graphical models \citep{DBLP:books/daglib/0066829}.
Finding the MAP configuration can easily be applied to finding the optimal joint action in a CG \citep{DBLP:conf/smc/VlassisEK04}.
In graphs without cycles, MP converges in a finite number of iterations~\citep{DBLP:books/daglib/0066829}. 
In these graphs, the right message passing ordering can result in convergence in the exact value and a speed up in convergence~\citep{DBLP:journals/spm/Loeliger04}. 
MP has no guaranteed converges on cyclic graphs, where the algorithm is synonymous with \textit{loopy belief propagation}~\citep{DBLP:conf/pkdd/KuyerWBV08}.
MP has a run-time complexity of $\mathcal{O}(|\mathcal{V}| \cdot |\mathcal{E}|)$ \citep{DBLP:journals/jair/ChoudhuryGMK22}. In our experiments, we use MP with its anytime extension and employ message-passing ordering identical to the elimination ordering of VE.
\begin{algorithm}[htbp]
    \caption{Max-Plus anytime action selection algorithm}
    \label{algo:mp}
    \begin{algorithmic}[1]
        \Require Coordination Graph $\cg = (\mathcal{V}, \mathcal{E})$, Edge Q-functions  $Q_{e}$ for $e \in \mathcal{E}$, $q = -\infty$
        \Procedure{MaxPlus}{$M$}
        \State $\mu_{ij}(a_j) = 0$ for $(i, j) \in \mathcal{E}$, $a_j \in \mathcal{A}_j$ 
        \For{$t \gets 1$ to $M$}
            \For{$i \in \mathcal{V}$}
                \For{$j \in \Gamma(i)$}
                    \State Compute $\mu_{ij}$ using \cref{eq:mpmsg} 
                    % \Comment{Message passing.}
                    \State $\mu_{ij}(a_j) \gets \mu_{ij}(a_j) - \frac{1}{|\mathcal{A}_j|} \sum_j \mu_{ij}(a_j)$ 
                    % \Comment{Message normalization.}
                \EndFor
                \State $\joint{a}_i \gets \arg\max_{a_i} \sum_{j \in \Gamma(i)} \mu_{ji}(a_i)$ 
            \EndFor
            \If{$\hat{Q}(\joint{a}) > q$} \Comment{Anytime extension.}
                \State $\joint{a}^* \gets \joint{a}$
                \State $q \gets u(\joint{a})$
            \EndIf
            \If{$|| \mu_{ij}^t - \mu_{ij}^{t-1} || \approx 0$} \Comment{Signal for convergence.}
                \State \textbf{break}
            \EndIf
        \EndFor
        \State \Return $\joint{a}^*$
        \EndProcedure
    \end{algorithmic}
\end{algorithm}
\noindent
Messages are sent between nodes in an iterative fashion. The messages $\mu_{ij}$ represent the running updates of the locally optimized payoff functions~\citep{DBLP:conf/robocup/KokV05} between agents $i$ and $j$, with $e = (i, j)$, over the edges $e \in \mathcal{E}$ of the graph $(\mathcal{V}, \mathcal{E})$. Messages are computed by the following equation:
\begin{equation}\label{eq:mpmsg}
    \mu_{ij}(a_j) = \max_{a_i} \left[Q_{ij}(a_i, a_j) + \sum_{k\in\Gamma(i)\setminus j} \mu_{ki} (a_i)\right],
\end{equation}
where $\Gamma(i)$ denotes the neighbors of $i$ in the graph.
In practice, the messages are normalized after the maximization by subtracting a normalization constant $c_{ij} = \frac{1}{|\mathcal{A}_j|} \sum_{a_j\in\mathcal{A}_j}\mu_{ij}(a_j)$ from \cref{eq:mpmsg} to prevent the messages from continuously increasing, which is especially prevalent in graphs with cycles \citep{DBLP:journals/sac/WainwrightJW04}. Messages are passed until they converge, or an anytime signal is received. The Max-Plus algorithm is anytime since messages are approximations of the exact value, meaning performance and computational efficiency can be interchanged. Furthermore, the algorithm allows for potentially increased decentralization, as messages need only to sum over the received messages from its neighbors defined over the actions of the receiving instance instead of enumerating all possible action combinations in VE.
We refer the interested reader to the extensive empirical comparison of \citet{DBLP:conf/robocup/KokV05} on the performance of MP and VE in coordination graphs with fewer confounding factors.

\section{Unweighted particle filtering in FT-POMCP} \label{appendix:ft-pomcp}
For FT-POMCP, we consider an unweighted filtering variant, similar to the weighted variant in FT-W-POMCP. 
In the original paper on FT-POMCP, 
\citet{DBLP:conf/aaai/AmatoO15} mention using separate particle filters for each edge in the factored trees algorithm, which is confirmed by the pseudo-code, but there are no details on the filtering procedure. The available information makes it unclear how root-sampling was executed in FT-POMCP. 
Therefore, we assumed that there are also multiple filters, one for each tree, which are sampled uniformly. We clarify the details of our implementation of the filters below.
\paragraph{Procedure.}
The offline \emph{unweighted} filter in FT-POMCP, $\overline{b}_e=\{(s_k)\}_{k=1}^{K_e}$, rejects based on the local observation $\overline{b}_e = \{s_k'\colon \joint{o}_e = \joint{o}^{\;(k)}_{e}\}$.
This procedure reflects how simulated particles $B(\joint{h}_e)$ would be added to the nodes of the trees that represent local history $\joint{h}_e$. 
Additionally, we merge the simulated particles $B(\joint{h}_e)$ from the tree with the offline belief $\overline{b}_e$ if few particles remain to decrease the chance of deprivation.

Thereby, the belief-state is approximated as:
$
b(s) \approx \sum_{e\in\mathcal{E}} \frac{1}{|\mathcal{E}|} \overline{b}_e(s),
$
where
$
\overline{b}_e(s) = \left(\frac{1}{K_e} \sum_{k=1}^{K_e} \delta(s, s^{(k)})\right)
$

\section{Pseudo-codes}\label{sec:pseudocode-appendix}

\subsection{Online Planning}
\cref{alg:op} summarizes the iterative online planning procedure as mentioned in the online planning paragraph of the main paper.
Planning starts from the initial belief $b_0$, which is assigned to the offline belief $\overline{b}$ maintained during the episode.
This belief is passed to the searching algorithm, as searching is only performed in the reachable states as given from the current belief.
The simulator $\mathcal{G}$ is called without a state to demonstrate that states are not assumed observable.
In practice, the episode is executed from some random initial state as defined by the environment, and this state is updated by the simulator sequentially.
Moreover, note that we leave the update of the trees implicit here for generality.
The computational budget is spent primarily in line 4 of \cref{alg:op}, where in our experimental evaluation, we restrict both the maximum time spent searching or the maximum number of \textsc{Simulate} calls, and the search is interrupted by violation of either constraint.
An example of an interrupted signal can originate from an end-of-episode signal or, for instance, by receiving a terminal state from the simulator.

% \begin{minipage}[t]{0.45\textwidth}
    \begin{algorithm}[ht]
    \caption{Generative step for a particle belief.}
    \label{alg:gpf}
    \begin{algorithmic}[1]
        \Procedure{$\mathcal{G}_{PF}$}{$\tilde{b}, \joint{a}$}
            \State $s_0 \sim \tilde{b}$
            \State $\joint{o} \gets \mathcal{G}(s_0, \joint{a})$
            \State $\tilde{b}' \gets \emptyset, C \gets |\tilde{b}|$
             \For{$k \gets 1 \text{ to } C$}
                \State ${s'}^{(k)}, r^{(k)} \gets \mathcal{G}(s^{(k)}, \joint{a})$
                \State ${w'}^{(k)} \gets w^{(k)}\mathcal{O}(\joint{o} \given {s'}^{(k)}, \joint{a})$
                \State $\tilde{b}' \gets \tilde{b}' \cup \{({s'}^{(k)}, {w'}^{(k)}\}$
            \EndFor
            \State $\rho \gets \tfrac{\sum_k w^{(k)} r^{(k)}}{\sum_k w^{(k)}}$
            \State $\tilde{b}' \gets \{{s'}^{(k)}, \tfrac{{w'}^{(k)}}{\sum_l {w'}^{(l)}})\}_{k=1}^C$ \Comment{Normalization.}
            \State \Return $\tilde{b}', \rho$
        \EndProcedure
    \end{algorithmic}
\end{algorithm}
% \end{minipage}
% \begin{minipage}[t]{0.45\textwidth}
    \begin{algorithm}[ht]
    \centering
    \caption{Online planning methodology}\label{alg:op}%
    \begin{algorithmic}[1]
        \Procedure{Execute}{$b_0$}
            \State $\overline{b} \gets b_0$
            \Repeat
                \State $\joint{a}^\# \gets \Call{Search}{\overline{b}}$
                \State $\joint{o}, r \gets \mathcal{G}(\cdot\given \joint{a}^\#)$ % \Comment{Execute step in real environment.}
                \State $\overline{b} \gets \Call{Update}{\overline{b}, \joint{a}^\#, \joint{o}}$ % \Comment{Update belief with $a^*$ and received $o$.}
            \Until $\Call{Interrupted}{}$
        \EndProcedure
    \end{algorithmic}
\end{algorithm}
% \end{minipage}

\subsection{Weighted Particle Filtering}\label{app:wpf}
In \cref{algo:sir}, we give the pseudo-code for our weighted particle filtering method used throughout the paper. The parameter $\tau$ is the threshold value when dividing the number of particles in the filter by the ESS, which indicates the need for re-sampling, which we set to $\tau=2$~ \citep{DBLP:journals/jstsp/SeptierP16,DBLP:conf/nips/WuYZYLLH21}.

To update each of the local beliefs $\overline{b}_e$ in FT-W-POMCP and FT-PFT as explained in \cref{sec:scalable_pf_method} of the main paper, we call the WPF procedure of \cref{algo:sir} with
WPF($\joint{a}$, \textbf{$\joint{o}_e$}, $\overline{b}_e$, $\mathcal{L}_e$, $\mathcal{O}_e$, $\mathcal{T}$, $\tau$) instead of WPF($\joint{a}, \joint{o}, \overline{b}$, $\mathcal{L}$, $\mathcal{O}$, $\mathcal{T}$, $\tau$). That is, with $\overline{b}_e, \joint{o}_e, \mathcal{O}_e, \mathcal{L}_e$ instead of $\overline{b}, \joint{o}, \mathcal{O}, \mathcal{L}$, respectively. Then, $\overline{b}_e$ is the local offline belief, $\joint{o}_e$ derived from $\joint{o}$, $\mathcal{O}_e$ the local observation probability, and $\mathcal{L}_e = \mathcal{L}(\overline{b}_e)$ is the likelihood of the local offline belief.

\begin{algorithm}[htbp]
    \caption{Weighted particle filtering.}
    \label{algo:sir}
    \begin{algorithmic}[1]
        \Procedure{WPF}{$\joint{a}, \joint{o}, \{(s^{(k)}, w^{(k)})\}_{k=1}^K$, $\mathcal{L}$, $\mathcal{O}$, $\mathcal{T}$, $\tau$}
        \State $w \gets 0$ \Comment{Keep track of total weight.}
        \State $\overline{b} \gets \emptyset$ \Comment{Initialize new intermediate belief.}
        \For{$k \gets 1 \text{ to } K$}
            \State ${s'}^{(k)} \sim \mathcal{T}(\cdot \given {s'}^{(k)}, \joint{a})$
            \State ${w'}^{(k)} \gets w^{(k)} \cdot \mathcal{O}(\joint{o} \given {s'}^{(k)}, \joint{a})$
            \State $\overline{b} \gets \overline{b} \cup \{\tuple{{s'}^{(k)}, {w'}^{(k)}}\}$
            \State $w \gets w + {w'}^{(k)}$
        \EndFor
        \State $\overline{b} \gets \{\tuple{{s'}^{(k)}, \tfrac{{w'}^{(k)}}{w}}\}_{k=1}^K$ \Comment{Normalization.}
        \If{$\tfrac{|\overline{b}|}{ESS(\overline{b})} \leq \tau$} \Comment{ESS threshold.}
            \State $\overline{b}'\gets\overline{b}$\Comment{Insignificant weight disparity.}
        \Else
            \State $\overline{b}' \gets \{\tuple{{s''}^{(k)}, \tfrac{1}{K}}\} \sim \overline{b}$ \Comment{Re-sampling.}
        \EndIf
        \State $\mathcal{L}' \gets \mathcal{L} \cdot w$
        \State \Return $\overline{b}', \mathcal{L}'$
        \EndProcedure
    \end{algorithmic}
\end{algorithm}

\subsection{FT-POMCP \& FT-W-POMCP}
FT-POMCP was introduced by \citet{DBLP:conf/aaai/AmatoO15}, which included the pseudo-code of the \textsc{Simulate} procedure in the supplementary material.
Our definition largely follows theirs but differs slightly in how we determine a \textsc{Rollout} in the pseudo-code, as it is unclear what exactly decides a \textsc{Rollout} in their definition.
In the current form in the appendix of \citet{DBLP:conf/aaai/AmatoO15}, a \textsc{Rollout} is always executed, which is, presumably, not the intended behavior of the procedure.
We show our pseudo-code here for completeness. Note that the pseudo-code also describes FT-W-POMCP, but we do not have \cref{code:ft-pomcp}.
\begin{algorithm}[htbp]
    \caption{FT-POMCP's \textsc{Simulate} procedure.}
    \label{algo:pomcpsim}
    \begin{algorithmic}[1]
        \Procedure{Simulate}{$s, h, d$}
            \If{$\gamma^d < \epsilon$}
                \State \Return $0$ %\Comment{Return if the maximum depth is reached.}
            \EndIf
            \ForAll{$e \in \mathcal{E}$} %\Comment{Enumerate every component/factor tree.}
                \If{$\joint{h}_e \not \in T_e$} %\Comment{If tree does not contain history.}
                    \ForAll{$\joint{a}_e \in \vec{\mathcal{A}_e}$} % \Comment{Enumerate component actions.}
                        \State $T_e(\joint{h}_e, \joint{a}_e) {\gets} (N_0(\joint{h}_e, \joint{a}_e), V_0(\joint{h}_e, \joint{a}_e), \emptyset)$ % \Comment{Construct nodes.}
                    \EndFor
                \EndIf
            \EndFor
            \If{$\exists_{e\in \mathcal{E}} : \joint{h}_e \not \in T_e$}
                \State \Return $\Call{Rollout}{s, \joint{h}, d}$ % \Comment{Value estimate by rollout.}
            \EndIf
            \State $\joint{a} \gets \max_{\joint{a}^\#} \sum_{e\in \mathcal{E}} \ucb(Q_e(\joint{h}_e, \joint{a}_e), N(\joint{h}_e), N(\joint{h}_e, \joint{a}_e))$ \Comment{VE or MP (\cref{eq:ucb}).}
            \State $\fgstep$ \Comment{Generative transition.}
            \State $R \gets r + \gamma\cdot{\Call{Simulate}{s', \joint{h}\joint{a}\joint{o}, d+1}}$
            \ForAll{$e \in \mathcal{E}$} \Comment{Update statistics for every tree.}
                \State $B(\joint{h}_e) \gets B(\joint{h}_e) \cup \{s\}$ \label{code:ft-pomcp} \Comment{Add state to local belief.}
                \State $N(\joint{h}_e) \gets N(\joint{h}_e) + 1$
                \State $N(\joint{h}_e, \joint{a}_e) \gets N(\joint{h}_e, \joint{a}_e) + 1$
                \State $Q_e(\joint{h}_e, \joint{a}_e) \gets Q_e(\joint{h}_e, \joint{a}_e) + \cfrac{R - Q_e(\joint{h}_e, \joint{a}_e)}{N(\joint{h}_e, \joint{a}_e)}$ % \Comment{Update expert.}
            \EndFor
            \State \Return $R$
        \EndProcedure
    \end{algorithmic}
\end{algorithm}
\subsection{FT-PFT}
As mentioned in \cref{sec:factored_pft}, the simulator needs to be extended to accept a particle-filter belief instead of a state. 
This extended simulator is depicted in \cref{alg:gpf}.
This algorithm follows the definition from \cite{lim2023optimality}.
The pseudo-code for the factored trees variant of Sparse-PFT is given in \cref{algo:ftpftsim}.
Here, $Ch_e(\tilde{b}, \joint{a}_e)$ indicates the list of children of tree node $T_e(\tilde{b}, \joint{a}_e)$ for every tree built for edge $e \in \mathcal{E}$ specifically.

\begin{algorithm}[htbp]
    \caption{FT-PFT's \textsc{Simulate} procedure.}
    \label{algo:ftpftsim}
    \begin{algorithmic}[1]
        \Procedure{Simulate}{$\tilde{b}, d$}
            \If{$\gamma^d < \epsilon$}
                \State \Return $0$ %\Comment{Return if the maximum depth is reached.}
            \EndIf
            \ForAll{$e \in \mathcal{E}$} %\Comment{Enumerate every component/factor tree.}
                \If{$\tilde{b} \not \in T_e$} %\Comment{If tree does not contain history.}
                    \ForAll{$\joint{a}_e \in \vec{\mathcal{A}_e}$} % \Comment{Enumerate component actions.}
                        \State $T_e(\tilde{b}, \joint{a}_e) {\gets} (N_0(\tilde{b}, \joint{a}_e), V_0(\tilde{b}, \joint{a}_e), \emptyset)$ % \Comment{Construct nodes.}
                    \EndFor
                \EndIf
            \EndFor
            \State $\joint{a} \gets \argmax_{\joint{a}^\#} \sum_{e\in \mathcal{E}} \ucb(Q_e(\tilde{b}, \joint{a}_e), N_e(\tilde{b}), N_e(\tilde{b}, \joint{a}_e))$ \Comment{VE or MP (\cref{eq:ucb}).}
            \If{$\exists_{e\in\mathcal{E}} \colon |Ch_e(\tilde{b}, \joint{a}_e)| < C$} 
            % \Comment{If any node can be expanded.}
                \State $\tilde{b}', \rho \gets \mathcal{G}_{PF}(\tilde{b}, \joint{a})$ \Comment{\cref{alg:gpf}.}
                \For{$e \in \mathcal{E}$}
                    \State $Ch_e(\tilde{b}, \joint{a}_e) \gets Ch_e(\tilde{b}, \joint{a}_e) \cup \{(\tilde{b}', \rho)\}$
                \EndFor
                \State $P \gets \rho + \gamma\cdot{\Call{Rollout}{\tilde{b}', d+1}}$
            \Else
                \State $\tilde{b}', \rho \sim \{Ch_e(\tilde{b}, \joint{a}_e) \mid e \in \mathcal{E}\}$
                \State $P \gets \rho + \gamma\cdot{\Call{Simulate}{\tilde{b}', d+1}}$
            \EndIf
            \ForAll{$e \in \mathcal{E}$} \Comment{Update statistics for every tree.}
                \State $N_e(\tilde{b}) \gets N_e(\tilde{b}) + 1$
                \State $N_e(\tilde{b}, \joint{a}_e) \gets N(\tilde{b}, \joint{a}_e) + 1$
                \State $Q_e(\tilde{b}, \joint{a}_e) \gets Q_e(\tilde{b}, \joint{a}_e) + \cfrac{P - Q_e(\tilde{b}, \joint{a}_e)}{N_e(\tilde{b}, \joint{a}_e)}$ % \Comment{Update expert.}
            \EndFor
            \State \Return $P$
        \EndProcedure
    \end{algorithmic}
\end{algorithm}

\section{Benchmark descriptions}\label{appendix:benchmarks}
In this section, we detail the benchmarks used in the empirical evaluation.
\subsection{Firefighting in a graph}
In a similar fashion to the introduction of the factored POMCP algorithms \citep{DBLP:conf/aaai/AmatoO15}, we adopt the \textsc{FireFightingGraph} (FFG) environment \citep{DBLP:conf/atal/OliehoekSWV08}. Agents are standing in a line, and houses are located to the left and right of each agent. Given $n$ agents, this gives $n_h = n+1$ houses in total. 
Each house has an associated fire level in $[0, n_f)$, with $n_f=3$, which indicates the severity of the fire. A state is comprised of the fire levels of each house, giving a state space of $|\mathcal{S}| = n_f^{n_h}$. 
Fire spreads more quickly between adjacent burning houses. The agents can move to either the left or right house in a deterministic fashion and receive a binary ``flames'' observation that indicates noisily whether the house the agent visited was burning.\footnote{Note that \citet{DBLP:conf/aaai/AmatoO15} introduced a third non-specified action, which we leave out.} 
The state, action $|\mathcal{A}|= 2^n$, and observation space $|\Omega|= 2^n$ grow exponentially with the number of agents. The rollout policy is a uniform random policy over the joint actions. 
Unless specified otherwise, we limit the maximum number of steps in the environment to $H=10$.

\subsection{Multi-agent rock sampling}
\textsc{RockSample} \citep{DBLP:conf/uai/SmithS04} is a traditional benchmark for online planners \citep{DBLP:conf/nips/SilverV10}. Similarly to \citet{DBLP:journals/ijrr/CaiLHL21}, we introduce Multi-Agent \textsc{RockSample} (MARS) by increasing the number of agents that need to be controlled. Particularly, we model the problem as an MPOMDP.
MARS environments are defined by their size $m$, the number of agents $n$, and the number of rocks $k$. 
The agents are initially spread out on the left-most x-coordinate on the $m\times m$ grid and are tasked with sampling rocks before leaving the area on the right-most x-coordinate. The rocks, good or bad with equal probability, are placed randomly across the grid but are not allowed to be placed at the initial positions of the robots. The action space is $|\mathcal{A}| = (5 + k)^n$, which consists of four compass actions that change the position of the robot deterministically, a sample action, and $k$ sensing actions, where $k$ is the number of rocks of the configuration. The observation space $|\Omega|= 3^n$, consisting of an uninformative observation and a noisy `good' and `bad' rock observation for sensing, is relatively small. 
The sensor noise decreases exponentially based on the Euclidean distance to the rock that is sensed and the efficiency of the sensor $\eta\in[0,1]$, where $\eta=0$ implies observations are uniform and $\eta=1$ implies perfect sensing \citep{DBLP:conf/uai/SmithS04}.
In our experiments, we set this parameter as $\eta=0.2$.
The rollout policy performs a sample action if a rover is located at a good rock and picks uniform random over the joint actions otherwise.
The agents receive a reward of $\pm10$ depending on whether the rock they sampled was good and $+10$ upon leaving the area on the right side.
During the search, there is a $-100$ penalty for leaving the grid at any other side (i.e., west, north, or south) of the map.
This $-100$ penalty can be found in the code of POMCP for the single-agent \textsc{RockSample} problem \citep{DBLP:conf/nips/SilverV10} and in HyP-DESPOT for MARS \citep{DBLP:journals/ijrr/CaiLHL21}, but it is not mentioned in either of their benchmark descriptions or in the original introduction of \textsc{RockSample} \citep{DBLP:conf/uai/SmithS04}.
Therefore, we reduce this penalty during episode execution to $-1$ to ensure results remain legible. 
Unless specified otherwise, we limit the maximum number of steps in the environment to $H=40$.

\begin{figure*}[htb]
    \centering
    \begin{tikzpicture}
    [
        level 1/.style = {black, sibling distance = 3cm, level distance = 1.5cm},
        level 2/.style = {black, sibling distance = 1.75cm},
        level 3/.style = {black, level distance = 1cm},
        % edge from parent path =  {(\tikzparentnode\tikzparentanchor) .. controls +(0,-1) and +(0,1) .. (\tikzchildnode\tikzchildanchor)}
    ]
     
    \node[draw, circle] (root){}%$\joint{h}^{t}_{e_{1}}$}
        child {node[draw, rectangle] {$\joint{a}_{e_1}^{\,1}$}
        child {node [draw, circle] {$\joint{o}_{e_1}^{\,1}$}}
        child {node [draw, circle] {$\joint{o}_{e_1}^{\,2}$}}
        edge from parent } 
        child {node[draw, rectangle] {$\vec{a}_{e_1}^{\,2}$}
        child {node [draw, circle] {$\joint{o}_{e_1}^{\,1}$}}
        child {node [draw, circle] {$\joint{o}_{e_1}^{\,2}$}}
        edge from parent };

    \end{tikzpicture}
    % \qquad
    \hfill
    \begin{tikzpicture}[baseline={(root.base)}]
        \node[text width=2cm] (e) at (0,3.25) {$\cdots \; e \; \cdots$};
    \end{tikzpicture}
    \hfill
    \begin{tikzpicture}
        [
        level 1/.style = {black, sibling distance = 3cm, level distance = 1.5cm},
        level 2/.style = {black, sibling distance = 1.75cm},
        level 3/.style = {black, level distance = 1cm},
        % edge from parent path =  {(\tikzparentnode\tikzparentanchor) .. controls +(0,-1) and +(0,1) .. (\tikzchildnode\tikzchildanchor)}
        ]
    \node[draw, circle] (root){}%$\joint{h}^{t}_{e_{|\mathcal{E}|}}$}
        child {node[draw, rectangle] {$\joint{a}_{e_{|\mathcal{E}|}}^{\,1}$}
        child {node [draw, circle] {$\joint{o}_{e_{|\mathcal{E}|}}^{\,1}$}}
        child {node [draw, circle] {$\joint{o}_{e_{|\mathcal{E}|}}^{\,2}$}}
        edge from parent } 
        child {node[draw, rectangle] {$\joint{a}_{e_{|\mathcal{E}|}}^{\,2}$}
        child {node [draw, circle] {$\joint{o}_{e_{|\mathcal{E}|}}^{\,1}$}}
        child {node [draw, circle] {$\joint{o}_{e_{|\mathcal{E}|}}^{\,2}$}}
        edge from parent };     
    \end{tikzpicture}
    \caption{FT-POMCP search trees.}
    \label{fig:fbt}
\end{figure*}
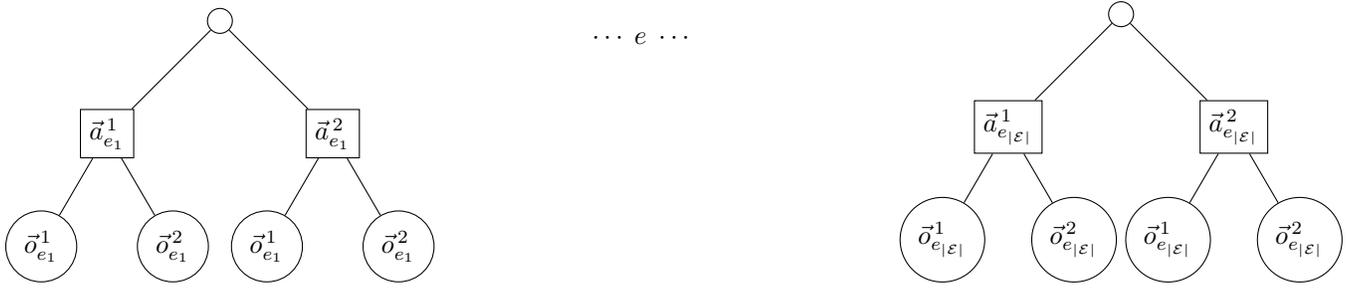

\subsection{Capture target (CT)}
Inspired by \citet{DBLP:conf/mrs/XiaoLA21}, we introduce \textsc{CaptureTarget} as an MPOMDP problem. $n$ agents move around and try to capture a single target in a  $m\times m$ grid world, with $m=12$. States consist of the $(x,y)$ coordinates of each agent and the target, giving a state space of $|\mathcal{S}| = m^{2n + 2}$. Actions consist of compass actions for each agent, giving $|\mathcal{A}| = 4^n$.
The target moves to the position which is the furthest away from all agents. Unlike the toroidal world of \citet{DBLP:conf/mrs/XiaoLA21}, our grid is confined by walls. 
Observations are binary and indicate whether the agent can `see' the target either horizontally or vertically in the grid, \ie, in Manhattan fashion. 
However, the observations are very noisy, as there is a 30\% chance of the target flickering, \ie, it can't be seen, and an additional 10\% sensor measurement imprecision for every agent. The observation space is thus $|\Omega| = 2^n$.
Rewards are very sparse; $+1$ if one or more agents capture the target and $0$ otherwise. 
CT is difficult due to the sparsity of the rewards. Therefore, the rollout policy is more involved as it picks the joint action that brings each agent closest to the target.
The episode ends if the target is captured or the maximum horizon is reached. 
Note that we report the averaged undiscounted rewards over $100$ episodes, which can be summarized as a \textit{capture rate}, \ie, the percentage of episodes in which the target was captured.
Unless specified otherwise, we limit the maximum number of steps in the environment to $H=50$.

        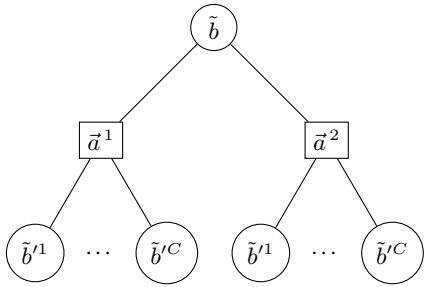
\begin{figure}[ht]\small
        \centering
        \begin{tikzpicture}
        [
            level 1/.style = {black, sibling distance = 3cm, level distance = 1.5cm},
            level 2/.style = {black, sibling distance = 1.75cm},
            level 3/.style = {black, level distance = 1cm},
            % edge from parent path =  {(\tikzparentnode\tikzparentanchor) .. controls +(0,-1) and +(0,1) .. (\tikzchildnode\tikzchildanchor)}
        ]
         
        \node[draw, circle] (root) {$\tilde{b}$} % $\tilde{b}^t$}
            child {node[draw, rectangle] (a1) {$\joint{a}^{\,1}$}
            child {node [draw, circle] (b11) {$\tilde{b}'^1$}}
            child {node [draw, circle] (b1c) {$\tilde{b}'^C$}}
            edge from parent } 
            child {node[draw, rectangle] (a2) {$\joint{a}^{\,2}$}
            child {node [draw, circle] (b21) {$\tilde{b}'^1$}}
            child {node [draw, circle] (b2c) {$\tilde{b}'^C$}}
            edge from parent 
            };
            \path (b11) -- node[auto=false]{\ldots} (b1c);
            \path (b21) -- node[auto=false]{\ldots} (b2c);
        \end{tikzpicture}
        \caption{Sparse-PFT tree.}
        \label{fig:particle_filter_tree}
        \end{figure}

\section{Trees for the variants of POMCP and Sparse-PFT}\label{appendix:trees}
This section contains the figures for the various belief trees mentioned in the main paper.
All (sets of) trees are depicted with one level of expansion depth.

% \begin{figure}[htbp]
    % \begin{minipage}{0.49\textwidth}

    % \end{minipage}%
    % \begin{minipage}{0.49\textwidth}
        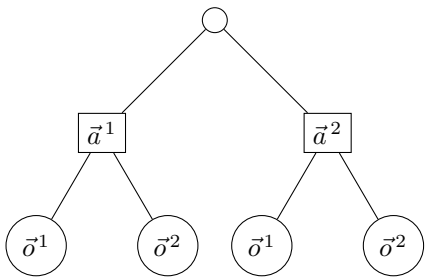
\begin{figure}[ht]
        \centering
        \begin{tikzpicture}
        [
            level 1/.style = {black, sibling distance = 3cm, level distance = 1.5cm},
            level 2/.style = {black, sibling distance = 1.75cm},
            level 3/.style = {black, level distance = 1cm},
            % edge from parent path =  {(\tikzparentnode\tikzparentanchor) .. controls +(0,-1) and +(0,1) .. (\tikzchildnode\tikzchildanchor)}
        ]
        % \small
        \node[draw, circle] (root) {}
            child {node[draw, rectangle] (a1) {$\joint{a}^{\,1}$}
            child {node [draw, circle] {$\joint{o}^{\,1}$} edge from parent node[left,draw=none]  {}
            }
            child {node [draw, circle] {$\joint{o}^{\,2}$} edge from parent node[right,draw=none]  {}
            }
            edge from parent } 
            child {node[draw, rectangle] (a2) {$\joint{a}^{\,2}$}
            child {node [draw, circle] {$\joint{o}^{\,1}$} edge from parent node[left,draw=none] {}
            }
            child {node [draw, circle] {$\joint{o}^{\,2}$} edge from parent node[right,draw=none] {}
            }
            edge from parent };
        \end{tikzpicture}
        \caption{POMCP search tree.}% for an MPOMDP with some history $\joint{h}$.}
        \label{fig:bt}
        \end{figure}
    % \end{minipage}
% \end{figure}

\begin{figure*}[htb]
    \centering
    \begin{tikzpicture}
    [
        level 1/.style = {black, sibling distance = 3cm, level distance = 1.5cm},
        level 2/.style = {black, sibling distance = 1.75cm},
        level 3/.style = {black, level distance = 1cm},
        % edge from parent path =  {(\tikzparentnode\tikzparentanchor) .. controls +(0,-1) and +(0,1) .. (\tikzchildnode\tikzchildanchor)}
    ]
     
    \node[draw, circle] (root) {$\tilde{b}$}%$\tilde{b}^t$}
        child {node[draw, rectangle] (a1) {$\joint{a}^{\,1}_{e_1}$}
        child {node [draw, circle] (b11) {$\tilde{b}'^1$}}
        child {node [draw, circle] (b1c) {$\tilde{b}'^C$}}
        edge from parent } 
        child {node[draw, rectangle] (a2) {$\joint{a}^{\,2}_{e_1}$}
        % child {node [] (b21) {$\ldots$}}
        child {node [draw, circle] (b21) {$\tilde{b}'^1$}}
        child {node [draw, circle] (b2c) {$\tilde{b}'^C$}}
        edge from parent 
        };
        \path (b11) -- node[auto=false]{\ldots} (b1c);
        \path (b21) -- node[auto=false]{\ldots} (b2c);
    \end{tikzpicture}
    % \qquad
    \hfill
    \begin{tikzpicture}[baseline={(root.base)}]
        \node[text width=2cm] (e) at (0,3.25) {$\cdots \; e \; \cdots$};
    \end{tikzpicture}
    \hfill
    % \qquad
    \begin{tikzpicture}
        [
        level 1/.style = {black, sibling distance = 3cm, level distance = 1.5cm},
        level 2/.style = {black, sibling distance = 1.75cm},
        level 3/.style = {black, level distance = 1cm},
        % edge from parent path =  {(\tikzparentnode\tikzparentanchor) .. controls +(0,-1) and +(0,1) .. (\tikzchildnode\tikzchildanchor)}
        ]
    \node[draw, circle] (root) {$\tilde{b}$}%$\tilde{b}^t$}
        child {node[draw, rectangle] (a1) {$\joint{a}^{\,1}_{e_{|\mathcal{E}|}}$}
        child {node [draw, circle] (b11) {$\tilde{b}'^1$}}
        child {node [draw, circle] (b1c) {$\tilde{b}'^C$}}
        edge from parent } 
        child {node[draw, rectangle] (a2) {$\joint{a}^{\,2}_{e_{|\mathcal{E}|}}$}
        child {node [draw, circle] (b21) {$\tilde{b}'^1$}}
        child {node [draw, circle] (b2c) {$\tilde{b}'^C$}}
        edge from parent 
        };
        \path (b11) -- node[auto=false]{\ldots} (b1c);
        \path (b21) -- node[auto=false]{\ldots} (b2c);
    \end{tikzpicture}
    \caption{FT-PFT trees.}
    % Only the first action $\joint{a}^{\,1}_{e_i}$ node is expanded for visibility.}
    \label{fig:ftpft}
\end{figure*}

\paragraph{Flat trees.}
\cref{fig:bt,fig:particle_filter_tree} show the trees as constructed by regular POMCP and Sparse-PFT, respectively.
In POMCP (\cref{fig:bt}), a sparse tree consisting of histories, \ie, paths of action-observation sequences, is built.
These sequences are gathered from the traces left by the states sampled at the root during their simulation.
Contrarily, a tree for Sparse-PFT (\cref{fig:particle_filter_tree}) represents a sparse particle-belief tree.
Tree nodes are constructed for actions and subsequent belief nodes, where each action node can have up to $C$ children.
More specifically, the belief nodes contain particles, representing a particle-belief state. Note that the FS variants (FS(-W)-POMCP and FS-PFT) employ the same tree structure as the flat variants but maintain factored statistics in the action nodes.

\paragraph{Factored trees.} \cref{fig:fbt} visualizes the trees built by the FT-POMCP variant. Similarly, \cref{fig:ftpft} shows how factored trees are constructed for Sparse-PFT in FT-PFT.
The main difference is that for FT-POMCP, both the action and observation space are factored.
Only the action space is factored in the tree for FT-PFT, as we do not explicitly branch on observations.

\fi

\end{document}